\documentclass[letterpaper]{article} 
\usepackage[submission]{aaai24}  
\usepackage{times}  
\usepackage{helvet}  
\usepackage{courier}  
\usepackage[hyphens]{url}  
\usepackage{graphicx} 
\usepackage{float} 
\usepackage{parskip}  
\usepackage{natbib}  
\usepackage{caption} 
\usepackage{algorithm}
\usepackage{algpseudocode}
\usepackage{subfig}
\usepackage{graphicx}
\usepackage{amsmath}
 
\usepackage{amsthm}
\usepackage{tabularx}

\usepackage{cite}
\usepackage{amsmath,amssymb,amsfonts}
\usepackage{textcomp}
\usepackage{xspace}
\usepackage{makecell}
\usepackage{booktabs,siunitx}
\usepackage{diagbox}
\usepackage{array, multirow}

\usepackage{newfloat}
\usepackage{listings}
\DeclareCaptionStyle{ruled}{labelfont=normalfont,labelsep=colon,strut=off} 
\floatstyle{ruled}
\newfloat{listing}{tb}{lst}{}
\floatname{listing}{Listing}
\title{Intriguing Equivalence Structures of the Embedding Space of Vision Transformers}

\author{
    Shaeke Salman\textsuperscript{\rm 1}, 
    Md Montasir Bin Shams\textsuperscript{\rm 1}, 
    Xiuwen Liu\textsuperscript{\rm 1}
}
\affiliations {
    \textsuperscript{\rm 1}Department of Computer Science, Florida State University, FL 32306, USA\\
    \{salman, liux\}@cs.fsu.edu, mshams@fsu.edu, 
}

\begin{document}

\maketitle

\begin{abstract}
Pre-trained large foundation models play a central role in the recent surge of artificial intelligence, resulting in fine-tuned models with remarkable abilities when measured on benchmark datasets, standard exams, and applications. Due to their inherent complexity, these models are not well understood. While small adversarial inputs to such models are well known, the structures of the representation space are not well characterized despite their fundamental importance. In this paper, 
using the vision transformers as an example due to the continuous nature of their input space, 
we show via analyses and systematic experiments that the representation space consists of large piecewise linear subspaces where there exist very different inputs sharing the same representations, and at the same time, local normal spaces where there are visually indistinguishable inputs having very different representations. 
The empirical results are further verified using the local directional estimations of the Lipschitz constants of the underlying models. 
Consequently, the resulting representations change the results of 
downstream models, and such models are subject to overgeneralization and with limited semantically meaningful generalization capability. 
\end{abstract}

\section{Introduction}
Built on large pre-trained foundation models~\cite{bommasani2022}, 
applications have exhibited unprecedented capabilities for a wide range of tasks, setting new state-of-the-art on benchmark datasets, acing standard exams, and passing professional exams~\cite{Gpt42023Open,Protein2022Brandes,Usmle2023Kung,main2021iot,islam2023unsupervised,emdad2023towards,Law2023Chatgpt}. 
Loosely speaking, applications have a relatively (very) small application-specific component, which is fine-tuned on top of the shared foundation models.
Therefore we focus on the foundation models and the outputs of such models, referred to as representations and also embeddings.
Transformers have become a hallmark component in models for many applications and have led to significant improvements in performance \cite{vaswani2023attention,dosovitskiy2021image,Devlin2018Bert,main2022iot,islam2023fusion,feng2023attention}, but there is no systematic study of the underlying embeddings in terms of fundamental characteristics. Given a representation of a model, to understand the generalization and overgeneralization, one must know the equivalence classes of inputs that share the same representation as the downstream applications will treat them the same. Similarly, knowing the characteristics of resulting embeddings of semantically equivalent inputs is also crucial: if these inputs can have very different representations, the models underlying all applications will have limited consistent generalization. 

It is well known that neural networks as classifiers exhibit an intriguing property in that they are subject to adversarial attacks: some small changes to an input could result in substantial changes in the classifier's  outputs~\cite{goodfellow2015explaining,szegedy2014intriguing,chakraborty2018adversarial,madry2019deep}. Conceptually speaking, those inputs are the ones that are close to the decision boundaries but near the given input; finding them leads to an optimization problem tied to the classifier and heuristic methods such as the fast gradient sign method and related variations, are often effective~\cite{goodfellow2015explaining,kurakin2017adversarial,Chen_2017,moosavidezfooli2016deepfool}.
However, these methods cannot be applied to studying the equivalences of the underlying representations given by the models. 

\begin{figure*}[ht]
  \centering
  \includegraphics[width=0.98\textwidth]
  {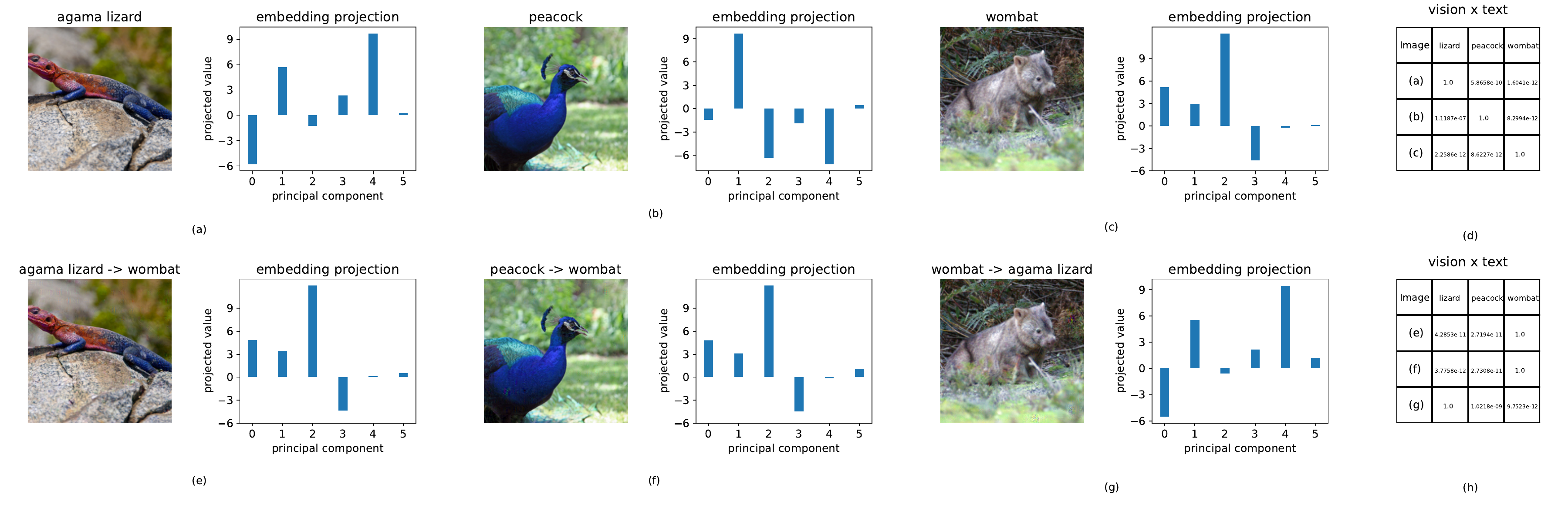}\label{fig:projection_embeddings}
  \vspace{-0.20in}
  \caption{Typical examples from ImageNet obtained using the proposed framework. 
   Three pairs of visually indistinguishable images (a and e, b and f, c and g) have different representations from each other as shown in their low-dimensional projections. In contrast, very similar representations are seen for the images in (e), (f), and (c), despite their substantial semantic differences; similar goes with images in (a) and (g). Note that the arrow in the title ($original \rightarrow target$) signifies a derived image from the original one by aligning the embedding of the original image with the target image using our method. The matrices (d) and (h) show the classification outcomes from the multimodal ImageBind pre-trained model used directly with no modifications. 
}
  \label{fig:overall}
\end{figure*}

\begin{figure}[ht]
    \centering
\begin{tabular}{>{\centering\arraybackslash}m{.08\textwidth}m{.35in}>{\centering\arraybackslash}m{.09\textwidth}m{.05in}>{\centering\arraybackslash}m{.1\textwidth}}
    \centering\arraybackslash
    \includegraphics[width=.10\textwidth]{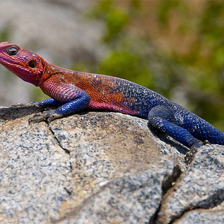} &%
    \centering\arraybackslash%
$\ +\ .02\ \times$ &%
    \includegraphics[width=.10\textwidth]{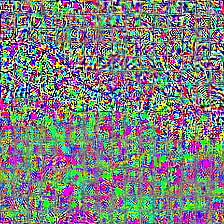} &%
    $=$ & %
    \includegraphics[width=.10\textwidth]{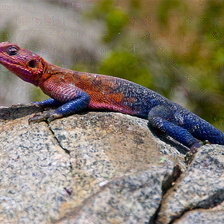} \\

    \centering\arraybackslash
    \includegraphics[width=.10\textwidth]{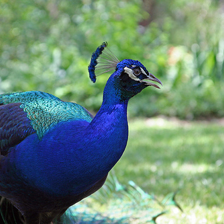} &%
    \centering\arraybackslash%
$\ +\ .02\ \times$ &%
    \includegraphics[width=.10\textwidth]{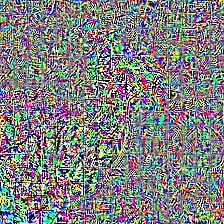} &%
    $=$ & %
    \includegraphics[width=.10\textwidth]{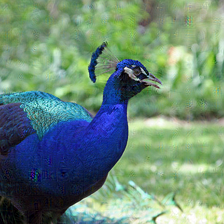} \\

    \centering\arraybackslash
    \includegraphics[width=.10\textwidth]{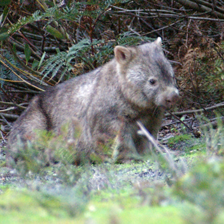} &%
    \centering\arraybackslash%
$\ +\ .02\ \times$ &%
    \includegraphics[width=.10\textwidth]{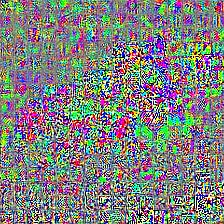} &%
    $=$ & %
    \includegraphics[width=.10\textwidth]{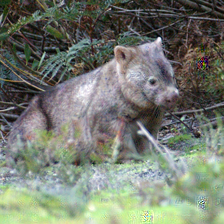} \\

\end{tabular}
    \caption{
  Pixel differences between the two images in each of the three pairs in Fig. \ref{fig:overall}; they are multiplied by 50 for visualization. 
    }
\label{fig:diffnoise}
\end{figure}

In this paper, using gradient-descent-based optimization procedures, we show empirically that perturbing an input to a deployed model in unnoticeable ways can alter the resulting representation to match that of any chosen one. Furthermore, we show that the resulting inputs will result in dramatic changes in classification results with no modifications to the classifiers. 
To highlight the key results of our framework, we use the ImageBind model as an example~\cite{girdhar2023imagebind}. Fig. \ref{fig:overall} shows several images along with their representations and the classification results. The three visually indistinguishable pairs in Fig. \ref{fig:overall}, (a) and (e), (b) and (f), and (c) and (g), respectively (see Fig. \ref{fig:diffnoise} for pixel differences) have very different representations, as shown by their low-dimensional projections.
On the other hand, the images in (e), (f), and (c) have very similar representations even though they are semantically very different; the images in (a) and (g) show another set.
When we pass these images to the unmodified multimodal ImageBind model, the images with similar embeddings are classified into the same class, regardless of their semantic similarity, as shown in Fig. \ref{fig:overall} (d) and (h).
These and additional results shown in the {\em Experimental Results} section, along with the fact we have obtained the same findings on all the images we have used, demonstrate convincingly that there are visually indistinguishable inputs having very different embeddings and yet that there are very different images having almost identical embeddings. Through the estimation of lower bounds on the local directional Lipschitz constants and the structures of the Jacobian matrices, we show such models are inherently vulnerable to adversarial attacks. Note that our method produces adversarial inputs as a by-product. 
By analyzing the equivalence classes of the embeddings of foundational models, the problem we solve is very different from the optimization problem for finding an adversarial input, and consequently, our results are more general and do not depend on application-specific classifiers.

Our main contributions are as follows:

\begin{itemize}
  \item We clearly demonstrate the algebraic and geometric structure of the embedding space of vision transformers. More specifically, we show that the input space consists of large piecewise linear subspaces where different images share the same representation and local normal spaces where visually indistinguishable images can have very different representations.
  \item We have proposed efficient computational procedures for finding equivalence structures of the embedding space and demonstrated their effectiveness in deployed models. As an additional outcome, we are able to identify adversarial examples to the representations which will affect all downstream applications.
  \item We show how to estimate the local directional Lipschitz constants robustly by understanding and overcoming the numerical issues of large models.
\end{itemize}



\section{Related Work}
With the availability of large datasets for challenging tasks such as natural language processing and computer vision, large foundational models have dominated the top-performing models and methods. In this new paradigm, such large models are trained on large datasets with huge computation, and then applications built on top with relatively small application-specific components to be further tuned using much smaller datasets. The trend has been further accelerated by the recent prompting-based models and multimodal models.
The joint multimodal models have demonstrated significant benefits by employing a shared embedding space across various modalities. One of them is ImageBind \cite{girdhar2023imagebind}, which aims to learn a single shared representation space by leveraging multiple types of image-paired data. 
This model aligns the embedding of each modality to image embeddings, resulting in an emergent alignment across all modalities via paired modeling based on the CLIP (the underlying vision and text model of ImageBind) \cite{radford2021learning}. The ImageBind results suggest semantic representation in the embedding space by matching embeddings of different modalities. 
Our results show that images with the same visual content can have highly dissimilar embeddings, whereas images with notable differences can have embeddings that are nearly identical. 

While recent works have improved the model performance on benchmark datasets and tasks, the fundamental issues of understanding how such models generalize, overgeneralize and memorize remain an open challenge~\cite{Zhang2016UnderstandingDL,zhang2017understanding,Neyshabur2017ExploringGI}.

Several researchers focus on the ``Activation Regions" concept and their potential role in understanding neural networks~\cite{crabbé2022concept}. 
Activation regions refer to specific regions in the input space that lead to certain activation patterns in the hidden layers of neural networks. 
Rectified Linear Units (ReLUs) are a common choice of activation functions in deep learning models. For neural networks where ReLU is used, they lead to piecewise linear regions~\cite{Montufar2014Onthe}. In their work, \citeauthor{Hanin2019deep}~(\citeyear{Hanin2019deep}) 
demonstrate that, despite the vast number of possible input patterns, deep neural networks with ReLU activations exhibit surprisingly few distinct activation patterns in their hidden layers. 
The linear approximation works well and we do not need to change an input much in order to match the embedding of another input.

Another line of research trying to understand the model is by probing the models to identify new properties. The most well-studied problem is adversarial attacks, where unnoticeable changes to the input can cause the models, mostly classifiers, to change their predictions.
\citeauthor{bhojanapalli2021understanding}~(\citeyear{bhojanapalli2021understanding}) and \citeauthor{shao2022adversarial}~(\citeyear{shao2022adversarial}) investigate the robustness of ViTs against attacks where the attacker has access to the model's internal structure. Their findings indicate that ViTs generally exhibit higher resilience than CNNs. 
\citeauthor{qin2023understanding}~(\citeyear{qin2023understanding}) and \citeauthor{salman2021certified}~(\citeyear{salman2021certified}) examine the robustness of Vision Transformers (ViTs) by focusing on the architectural structure based on patches.~\citeauthor{herrmann2022pyramid}~(\citeyear{herrmann2022pyramid}) additionally develop a pyramid adversarial training approach incorporating augmentation techniques to enhance both the sanity and robust performance of Vision Transformers (ViTs).
A recent work \cite{carlini2023aligned} explores the interplay between alignment techniques and adversarial attacks in neural networks, highlighting the potential vulnerabilities of aligned models. Though most adversarial examples have been applied to image classification tasks~\cite{szegedy2014intriguing}, the availability of multimodal models facilitates the application to text and other domains. In our work, based on a gradient-descent-based optimization procedure, we are able to find adversarial attacks to the embedding of any given image.

A common explanation of the existence of adversarial attacks is that the Lipschitz constants for deep neural networks are large, and therefore, models are sensitive to small changes \cite{fazlyab2023efficient,szegedy2014intriguing,goodfellow2015explaining}. Several papers focus on estimating the Lipschitz constants, both global and local.
Prior research works have shown that Lipschitz properties reveal intriguing behaviors of neural networks, such as robustness and generalization~\cite{szegedy2014intriguing}.
In recent times, numerous studies have delved into the exploration of optimization-based methods for bounding or approximating the Lipschitz constant of neural networks~\cite{scaman2019lipschitz,latorre2020lipschitz,fazlyab2023efficient}. ~\citeauthor{avant2021analytical}~(\citeyear{avant2021analytical}) determines guaranteed upper
bounds on the local Lipschitz constant of  larger neural networks
with ReLU activations.
The {LipsFormer} architecture by \citeauthor{qi2023lipsformer}~(\citeyear{qi2023lipsformer}), attempts to address the issue of training instability in transformers, a challenge particularly pronounced during the initial training phases. 
They derive theoretical upper limits for the Lipschitz constants, providing valuable insights into this aspect. Our results
are complementary in nature; we show the distributions
of the local directional Lipschitz constants of real trained
large models and are able to estimate them accurately.

\section{Preliminaries}

Understanding the large foundational models requires an understanding of all the components. However, such models are very complex due to the number of parameters used. To overcome the challenges, we roughly divide a model into two stages: a large foundational stage that is common to different applications and then an application-specific stage, consisting of classifiers and other application-specific components. To simplify the analyses, we assume the foundational model stage is fixed. 
As we focus on vision transformers, here we first describe the transformers mathematically and describe the vision transformers.

Transformers can be described mathematically succinctly, consisting of a stack of transformer blocks.
A {transformer block} is a parameterized function class $f_\theta: \mathbb{R}^{n \times d} \rightarrow \mathbb{R}^{n \times d}$. If $\mathbf{x} \in \mathbb{R}^{n \times d}$ then $f_\theta(\mathbf{x}) = \mathbf{z}$ where
$Q^{\left(h\right)}\left(\mathbf{x_i}\right) = W^T_{h,q}\mathbf{x}_i,\quad
    K^{\left(h\right)}\left(\mathbf{x_i}\right) = W^T_{h,k}\mathbf{x}_i,\quad
    V^{\left(h\right)}\left(\mathbf{x_i}\right) = W^T_{h,v}\mathbf{x}_i,\quad
    W_{h,q}, W_{h,k}, W_{h,v} \in \mathbb{R}^{d \times k}$.
The key multi-head self-attention is a softmax function applying row-wise on the inner products.\footnote{Note that there are other ways to compute the attention weights.}
\begin{equation}
    \alpha_{i,j}^{\left(h\right)} = \texttt{softmax}_j\left(\frac{\left<Q^{\left(h\right)}\left(\mathbf{x}_i\right),K^{\left(h\right)}\left(\mathbf{x}_j\right)\right>}{\sqrt{k}}\right)
\end{equation}
The outputs from the softmax are used as weights to compute new features, emphasizing the ones with higher weights given by
\begin{equation}
    \mathbf{u}'_i = \sum\limits_{h=1}^H W^T_{c,h} \sum\limits_{j=1}^n \alpha_{i,j} V^{\left(h\right)} \left(\mathbf{x}_j\right),\quad
    W_{c,h} \in \mathbb{R}^{k \times d}.
\end{equation}

The new features then pass through a layer normalization, followed by a ReLU layer, and then another layer normalization. 
Typically transformer layers are stacked to form deep models. Such models are used for natural language processing tasks, including various language models and machine translation. 

Recently the transformer architectures are adapted to vision tasks by using image blocks on the basic units, and spatial relationships among the units are captured via the self-attention mechanism. 
Since images can change smoothly and continuously, they make the analyses of the embedding space amendable to mathematical analyses. For example, vision transformers transform image patches into an embedding using a multi-layer perception applied on the output from the transformers~\cite{dosovitskiy2021image}.

While the proposed method applies to all transformer-based models with continuous inputs, we focus on the CLIP model~\cite{radford2021learning}, which jointly models images and text using the same shared embedding space used in the ImageBind model.

\section{Proposed Framework}
Here we describe the framework that enables us to explore the embedding space, analyze their properties, and verify them in large models. 
Generally, we model the representation given by a (deep) neural network (including a transformer) as a function $f: \mathbb{R}^m\rightarrow\mathbb{R}^n$. The fundamental question is to have a computationally efficient and effective way to explore the embeddings of inputs in the representation space by finding the inputs whose representation will match the one given by $f(x_{tg})$, where $x_{tg}$ is an input whose embedding we like to match. Informally, given an image of a lizard in Fig. \ref{fig:overall} as an example, all the images that share its representation given by a model will be treated as a lizard. 
In addition, we like to know the local algebraic and geometric structures of a representation; as adversarial examples are known to exist in neural network models as classifiers, we would like to know whether adversarial examples exist for representations. More importantly, we like to know how local spaces are connected.

\subsection{A Simple and Effective Procedure}
Note that it is much more challenging to find inputs that would match the representation of a target input. Since we need to match two vectors, we define the loss for finding an input matching a given representation as
\begin{equation}
   L(x) = L(x_0+\Delta x)= \frac{1}{2}\Vert f(x_0+\Delta x) - f(x_{tg})\Vert^2,
\end{equation}
where $x_0$ is an initial input and $f(x_{tg})$ specifies the target embedding. The gradient is given by
\begin{equation}
\frac{\partial L}{\partial x} \approx \left(\frac{\partial f}{\partial x}\Big|_{x=x_0}\right)^T(f(x_0+\Delta x) - f(x_{tg})).
\label{eq:grad_J}
\end{equation}
Eq. \ref{eq:grad_J} shows how the gradient of the mean square loss function is related to the Jacobian of the representation function at $x=x_0$. While optimal solutions can be obtained by solving a quadratic programming problem or linear programming problem, depending on the norm to be used when minimizing $\Delta x$, the gradient function works effectively for all the cases we have tested due to the Jacobian of the transformer.

One of the practical issues using the gradient descent-based procedure is how to determine the learning rate. In the case of the transformers, the model can be approximated by a linear model when it moves within one activation region; note that it is approximate due to the nonlinearity by the softmax, whose gradient is known. 
This property allows the gradient method to be very effective.
We call the procedure embedding matching procedure.

\subsection{Local Algebraic and Geometric Structures}
Given an input $x_0$, the local structures decide how the model behaves in the local neighborhood; for transformer-based models, note that the local neighborhood can be large spatially in the input space. Since we know the nonlinearity of the transformers is due to the ReLU function being used and the softmax function, the linear approximation of the function in a local neighborhood should be effective, given by
\begin{equation}
    f(x_0+\Delta x) \approx f(x_0) + \frac{\partial f}{\partial x}\Big|_{x=x_0} \times \Delta x.
\end{equation}
as in Eq. \ref{eq:grad_J}, $\frac{\partial f}{\partial x}$ is the Jacobian matrix of the function at $x=x_0$.
As a result, for deployed models, where $m>n$, there is a null space where the embeddings do not change as the input changes; it can be obtained via a reduced singular decomposition of the Jacobian. There is a normal space in the space perpendicular to the null space, where the embeddings can change quickly. 
To quantify how sensitive a representation is to local perturbations in the input space, we compute an accurate estimate of the extended local Lipschitz constant, given by the smallest $L$, such that 
\begin{equation}
    ||f(x_0+\Delta y)-f(x_0+\Delta x)|| \leq L ||\Delta y -  \Delta x||,
\end{equation}
where $\Delta x$ and $\Delta y$ specify the accepted neighborhood of $x_0$.
 Since the derivative of the ReLU function is not defined at 0, the definition avoids the issue. $L$ can be estimated accurately using the largest singular value of the Jacobian, and we also verify numerically.

As the model is high dimensional in nature, its behavior depends on the directions as well.
To quantify that, we also define and estimate the local directional Lipschitz constant ($L_{LDLC}$), along a given direction. The estimate is helpful to characterize how fast the model changes along the direction. Since $x_0$ could be near or even on the boundary between different activation regions in terms of the ReLU network, the $L_{LDLC}$ is defined as the smallest number that satisfies 
\begin{equation}
||f(x_0+\beta \Delta x_0)-f(x_0+\alpha \Delta x_0)|| \leq L_{LDLC} |\beta - \alpha|,
\end{equation}
where $0\leq |\alpha|, |\beta| \leq \epsilon_{LD}$, $\Delta x_0$ is a unit length vector, specifying the direction, and $\epsilon_{LD}$ is a parameter specifying the range of $\alpha$ and $\beta$. Estimated $L_{LDLC}$ values and their distributions allow us to quantify the changes in the normal space and the null space.

\subsection{Manifold Structures of the Embedding Space and Their Implications}
Putting all together, it is clear that the embedding space consists of subspaces  where the representations do not change locally and are therefore invariant to all the changes in the space; invariance to nuance changes is desirable, and results in generalization but invariance to other changes will lead to harmful overgeneralization. These subspaces together form a manifold in the space. Since ReLU is piecewise linear and reduces to a linear function within one activation region, the manifold is piecewise linear in nature, corresponding to the activation regions. The manifold is locally a subspace, and therefore the connection with the Grassmannian manifold can be exploited to characterize them formally~\cite{gallivan2003}. In this paper, we adopt a numerical approach and leave the formal exploration as future work.  There are also normal directions where the small changes in the input can lead to large changes in the representation, causing the model not to generalize well and be subject to adversarial attacks. The rate of change is bounded by the largest singular value of the Jacobian matrix and can be studied formally and numerically.

While the description is high level, we instantiate it using the CLIP model~\cite{radford2021learning}, a commonly deployed vision transformer. In addition, as the algebraic and geometric structures do not depend on the specifications of a model, we expect the results should be similar with other vision transformers and other models where the Jacobian can be estimated. We have validated this and provided detailed insights in the Experimental Results section and Appendix.

\section{Experiments}

In this section, we begin by providing the specifics of our experimental settings and implementation details. Our proposed framework is systematically applied across various datasets and multiple vision transformer models; in the subsequent subsections, we present both the experimental outcomes and quantitative results. 
\begin{figure}[H]
  \centering
  \includegraphics[width=0.9\columnwidth]
  {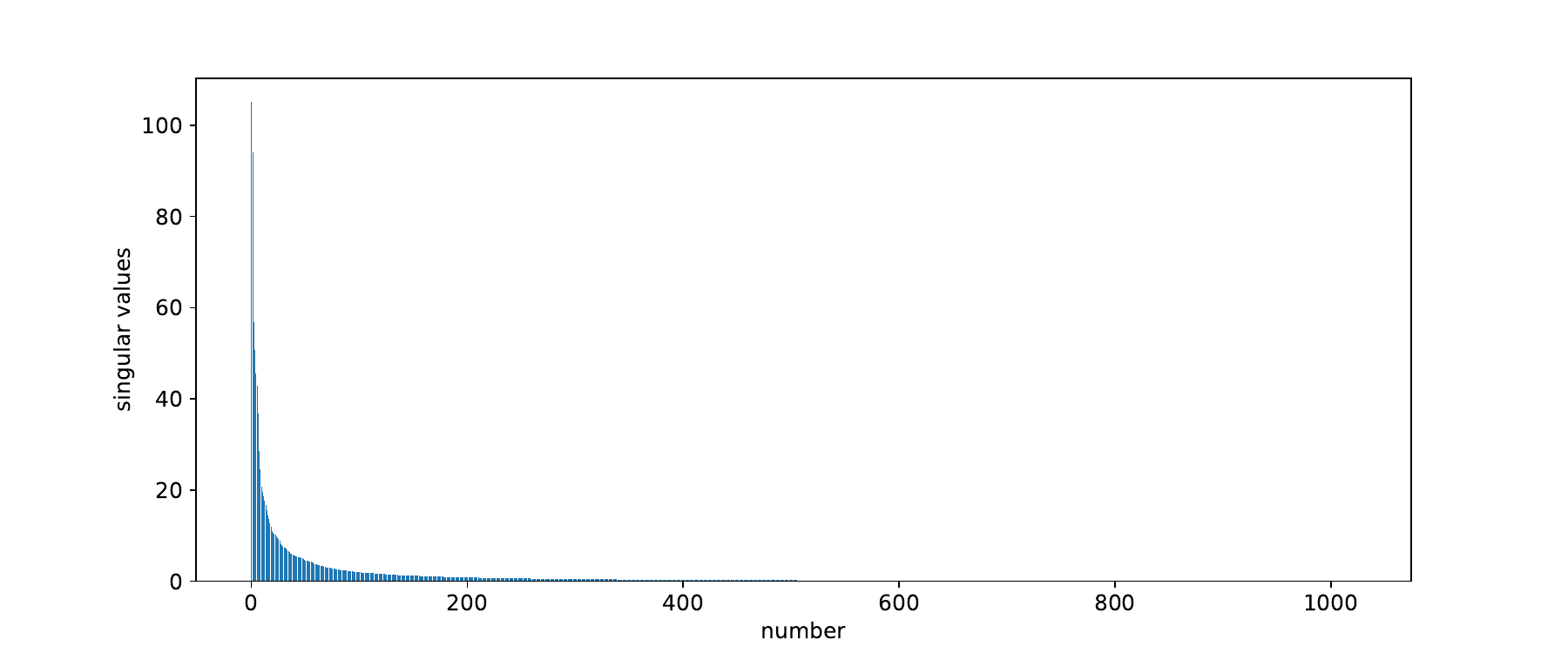} \\ 
  \includegraphics[width=0.9\columnwidth]
{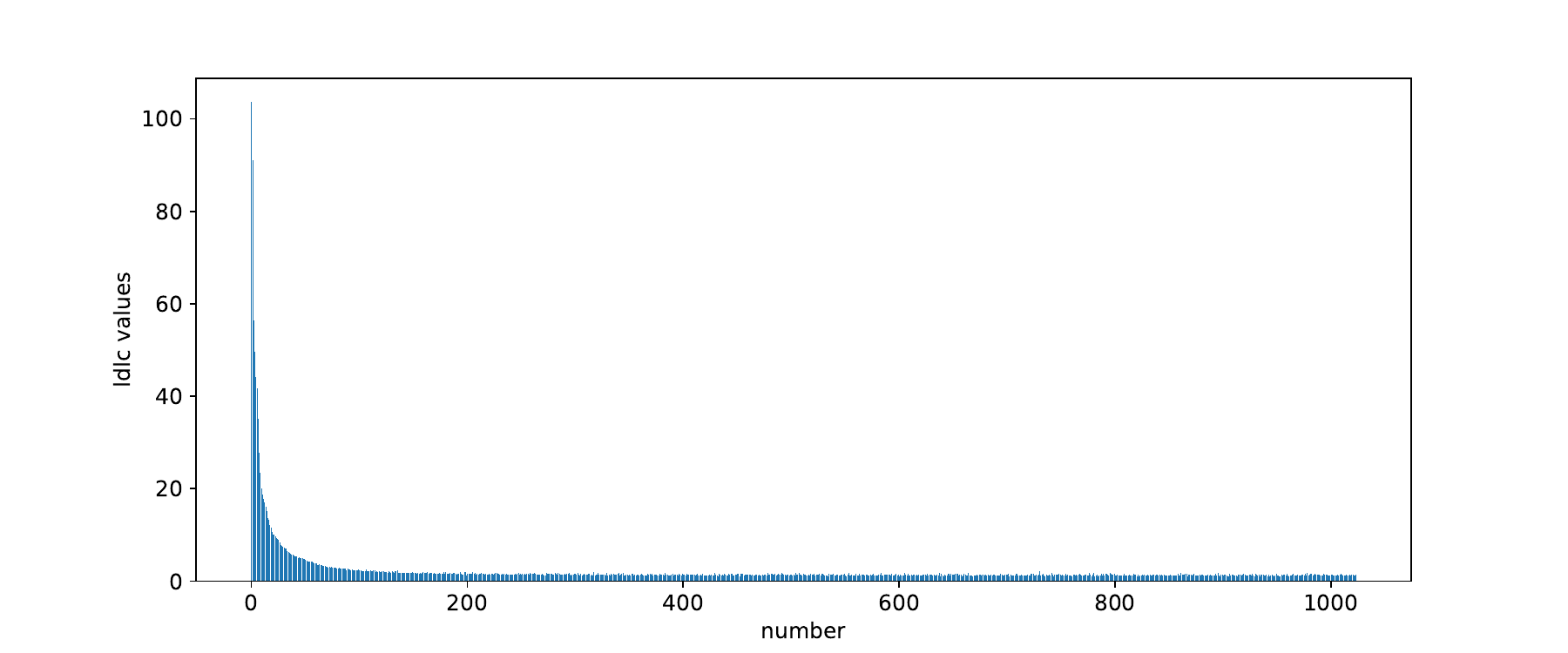}
  \vspace{-0.10in}
  \caption{Local structures of the embedding space. (top) The singular values of the Jacobian Matrix for Fig. \ref{fig:overall}(a); (bottom) The estimated local directional Lipschitz constant values along the directions given by the right singular vectors, which are consistent with the singular values.}
  \label{fig:singulars_lc}
\end{figure}

Our findings showcase the capability to align any image with another image through imperceptible adversarial attacks within a vision transformer model. More importantly, we show that our framework exhibits versatility, being agnostic to both the model architecture and dataset characteristics.

\subsection{Datasets and Settings}

\textbf{Datasets.} We conduct extensive experiments to evaluate our proposed framework on widely recognized vision datasets, namely ImageNet~\cite{imagenet2009Deng}, MS-COCO~\cite{lin2015microsoft} and Google Open Images~\cite{OpenImages}. 

\textbf{Implementation Details.} To demonstrate the feasibility of the proposed method on large models, we have used the pre-trained model publicly available by ImageBind\footnotemark\footnotetext{https://github.com/facebookresearch/ImageBind}, which in turn uses a CLIP model \footnotemark\footnotetext{https://github.com/mlfoundations/open\_clip}.
More specifically, ImageBind utilizes the pre-trained vision (ViT-H 630M params) and text encoders (302M params) from the OpenCLIP~\cite{ilharco_gabriel_2021_5143773,girdhar2023imagebind}. The input size is $224\times224\times3$, and the dimension of the embedding is 1024. As a result, the Jacobian matrix is of size $1,024\times 150,528$. 


\subsection{Experimental Results}

We have tested the embedding matching procedure using many image pairs.
Fig. \ref{fig:training_dynamics} shows a typical example, where the left one shows the evolution of loss when matching a specified target embedding. We use a small step size to make sure it converges. The right one shows that cosine similarity increases steadily. We also show the average pixel value difference between the new input and the original image at each step; one can see the values remain very small even though they increase as well. The algorithm is not sensitive to the learning rate and works effectively across a broad range of values, spanning from 0.001 to 0.09.  For instance, with a learning rate of 0.001, convergence is achieved in around 25,000 iterations, while 0.09 requires around 3,000 iterations. The visual differences in the resulting images are not noticeable. Eqn. 4 and 5 provide an explanation, as the gradient for our loss is insensitive to the learning rate. We will provide source code for all our experiments in GitHub\footnotemark\footnotetext{https://github.com/programminglove08/EquivalenceStruct}. 

\textbf{Quantitative evaluation.} We use reduced singular value decomposition to write the Jacobian as $U\Sigma V^T=\sum_{i=0}^{1023}s_i\times U(:, i)\times V(:, i)^T$,
where $T$ denotes the matrix transpose operator.
The top plot in Fig. \ref{fig:singulars_lc} shows the singular values of the Jacobian matrix in Fig. \ref{fig:overall}(a). The distribution of the singular values shows that the Jacobian has several dominating directions, reflecting the training set and the training algorithm being used. Note that the largest singular value gives us an estimation of the $L_{LDLC}$ at the input image. It shows that the model is sensitive to small changes along those directions. We also empirically estimate the $L_{LDLC}$values along the directions; the results are shown in the bottom plot of the figure. One can see that the values match well and indicate that the linear model provides a good approximation locally.

\begin{figure}[ht]
  \centering
  
  {\includegraphics[width=0.40\columnwidth]{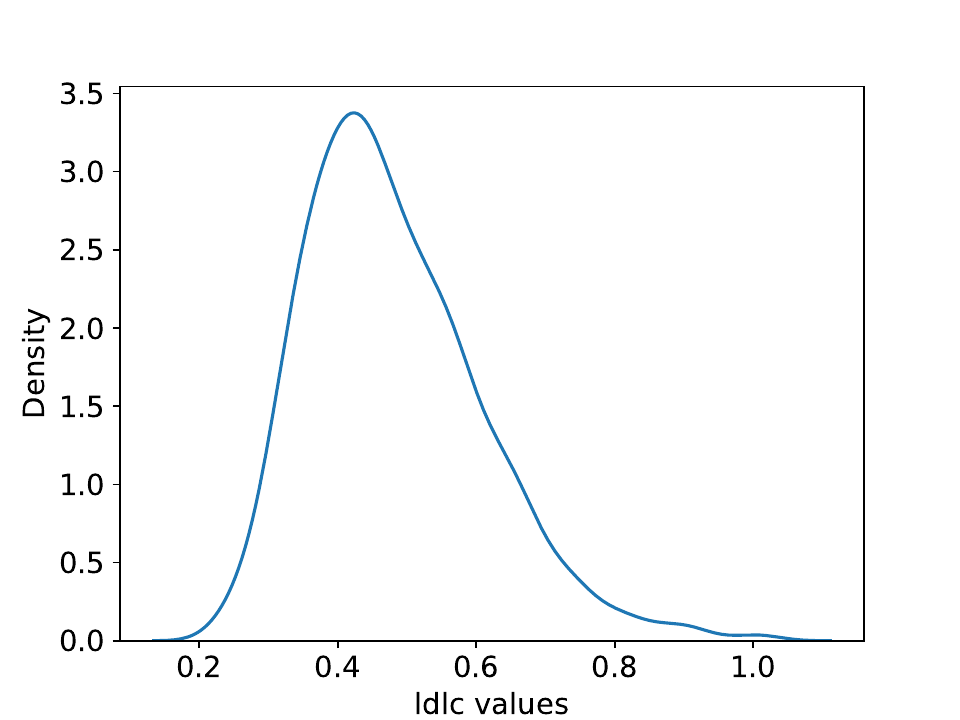}} 
  {\includegraphics[width=0.40\columnwidth]{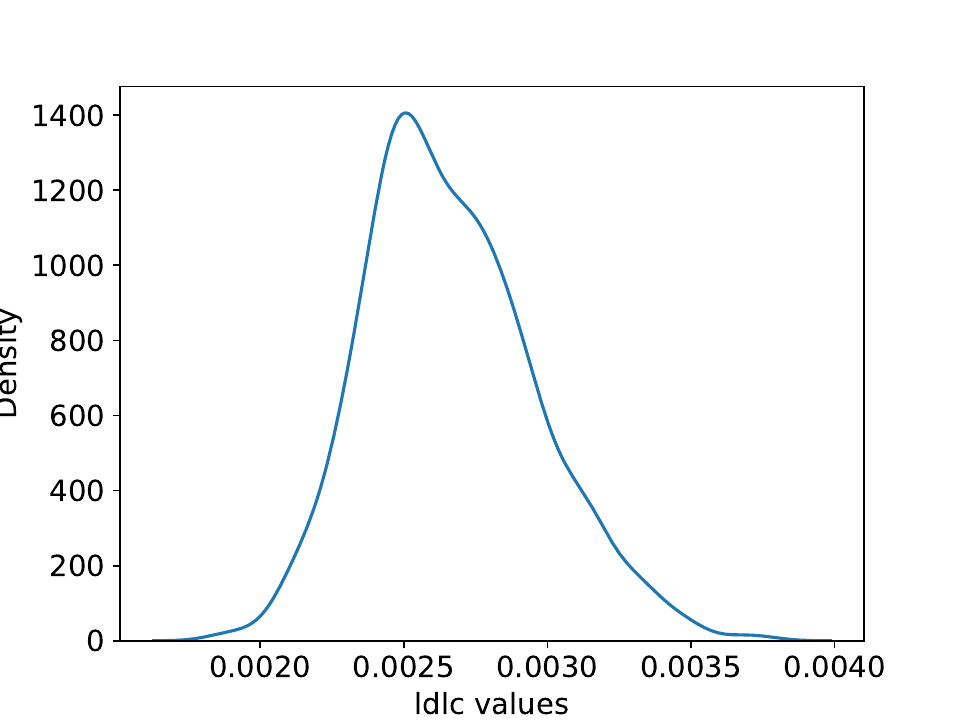}} \\
  {\includegraphics[width=0.94\columnwidth,height=0.3\columnwidth]{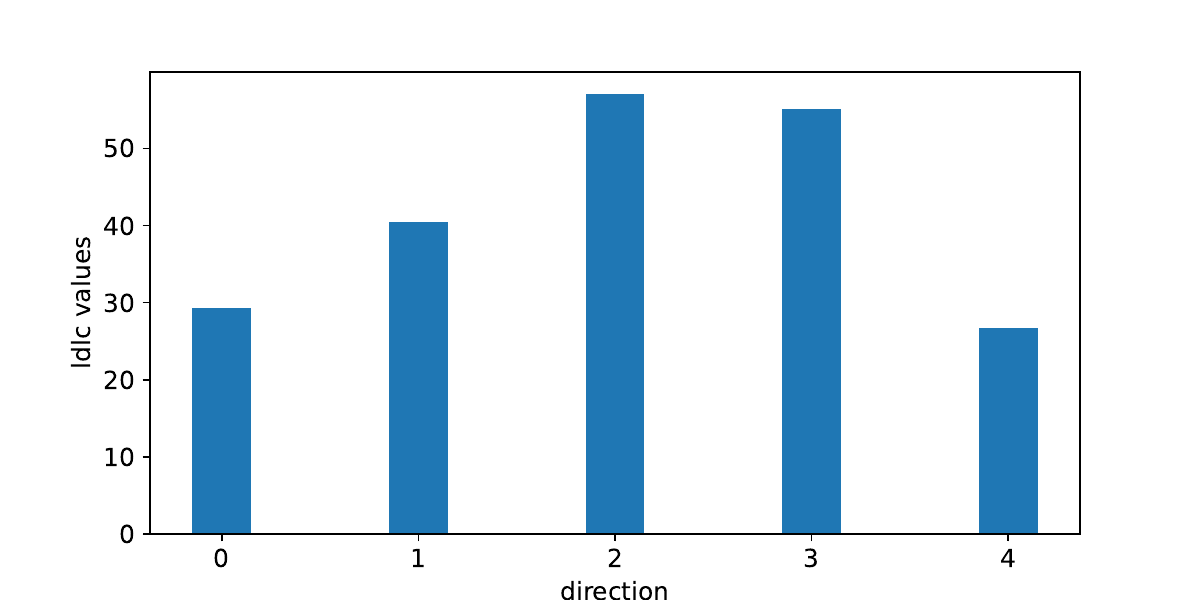}} 
 
  \vspace{-0.10in}
  \caption{The distribution of the estimated local directional Lipschitz constant values along the directions given by random Gaussian vectors (top left), random Gaussian vectors in the null space of singular vectors (top right), and the gradient optimization procedure (bottom). 
  }
  \label{fig:LC}
\end{figure}

\begin{figure}[ht]
  \centering
  {\includegraphics[width=0.2\textwidth]{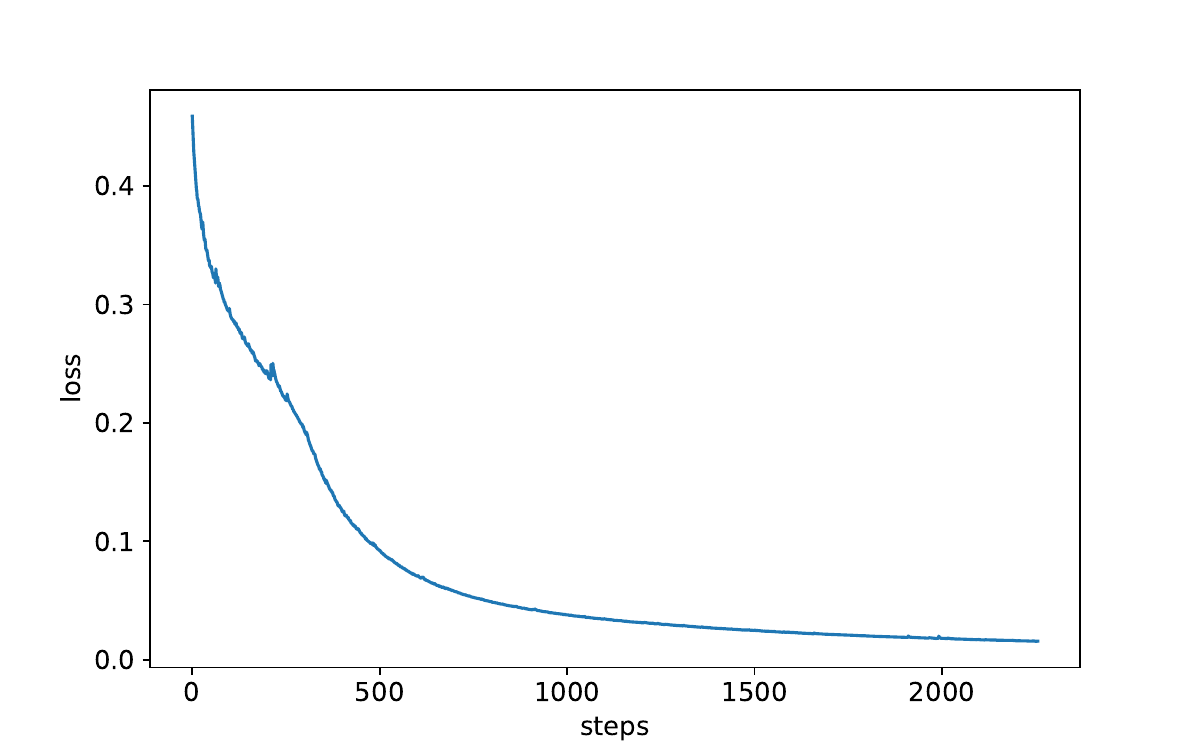}}
  {\includegraphics[width=0.2\textwidth]{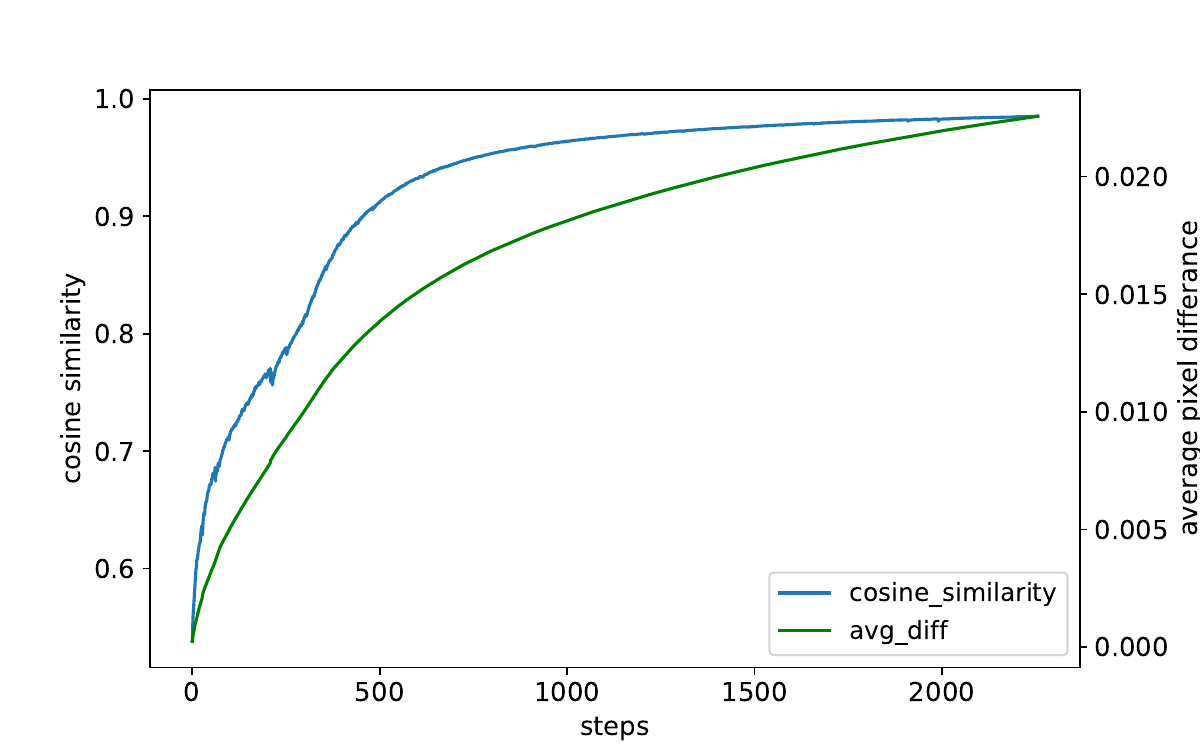}}
  \vspace{-0.10in}
  \caption{The evolution of loss while matching a target embedding. (left) the loss w.r.t. steps. (right) the cosine similarity between the embeddings of the new input and the target w.r.t. the steps, along with the average pixel value difference between the new input and the original image.}
  \label{fig:training_dynamics}
\end{figure}

\begin{figure}[ht]
  \centering
  \includegraphics[width=0.225\textwidth]{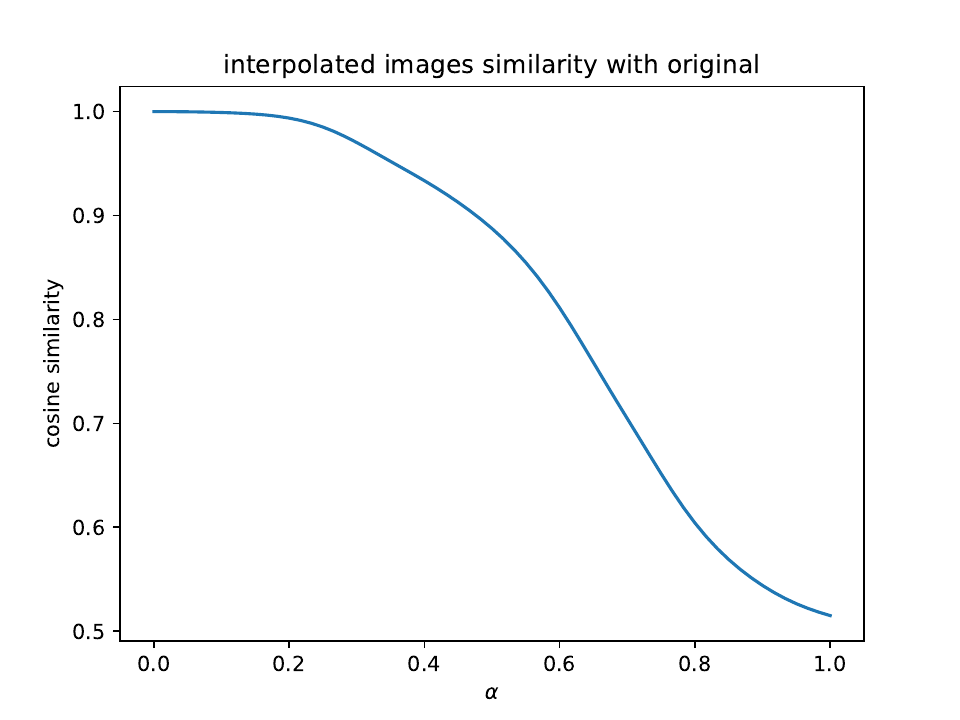}
  \includegraphics[width=0.225\textwidth]{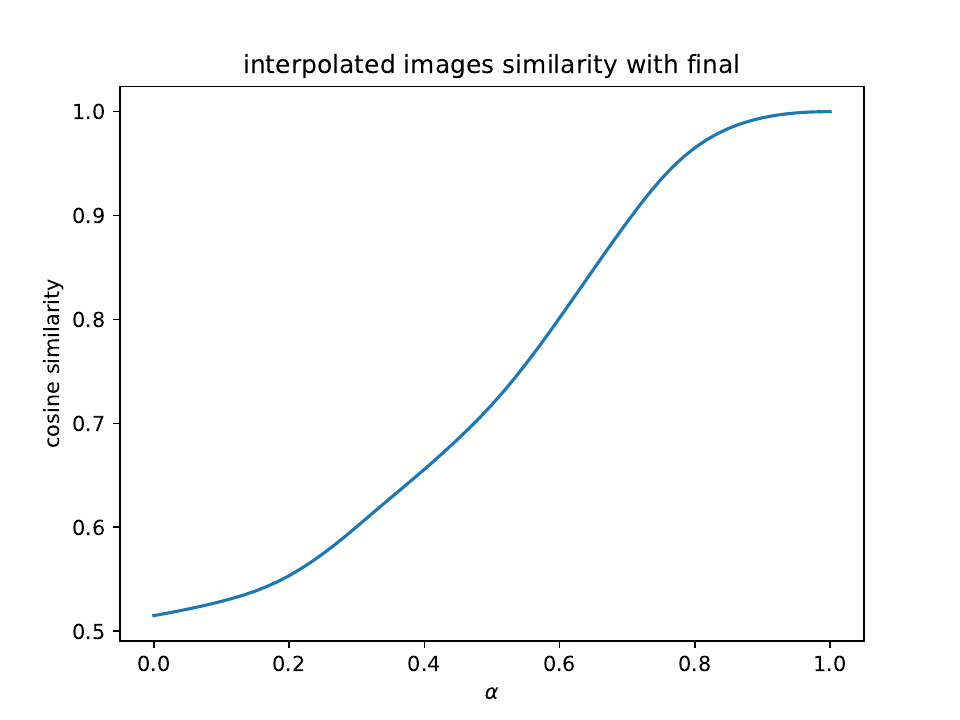}
  \vspace{-0.10in}
  \caption{The change of the embeddings as the input changes linearly for the image pair in Fig. \ref{fig:overall}(a). (left) cosine similarity between embeddings of the interpolated and the original. (right) same as left but for the final image.}
  \label{fig:nonlinearity}
\end{figure}

Fig. \ref{fig:LC} (top right) shows distributions of the estimated $L_{LDLC}$ values for 10,000 such randomly generated directions in the null space, and the values are consistently small. In comparison, Fig. \ref{fig:LC}  (top left) shows the same but for random directions. Note that when a random direction is used, the resulting direction is a mixture of the null subspace and normal space. As expected, their values are much larger than those in the null space. Fig. \ref{fig:LC} (bottom) shows the estimated $L_{LDLC}$ values along the directions given by our gradient optimization procedure. Those values are two orders of magnitude larger than the random directions and four orders of magnitude larger than the values in the null space, showing the effectiveness of the procedure. 

\begin{figure}[ht]
  \centering
  \includegraphics[width=0.225\textwidth]{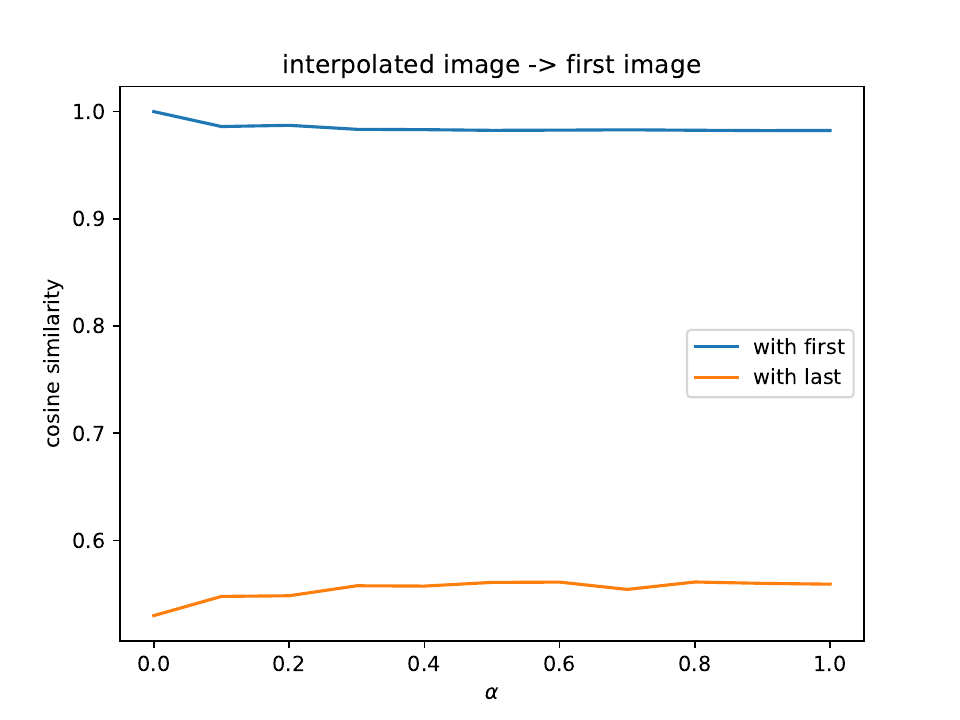}
  \includegraphics[width=0.225\textwidth]{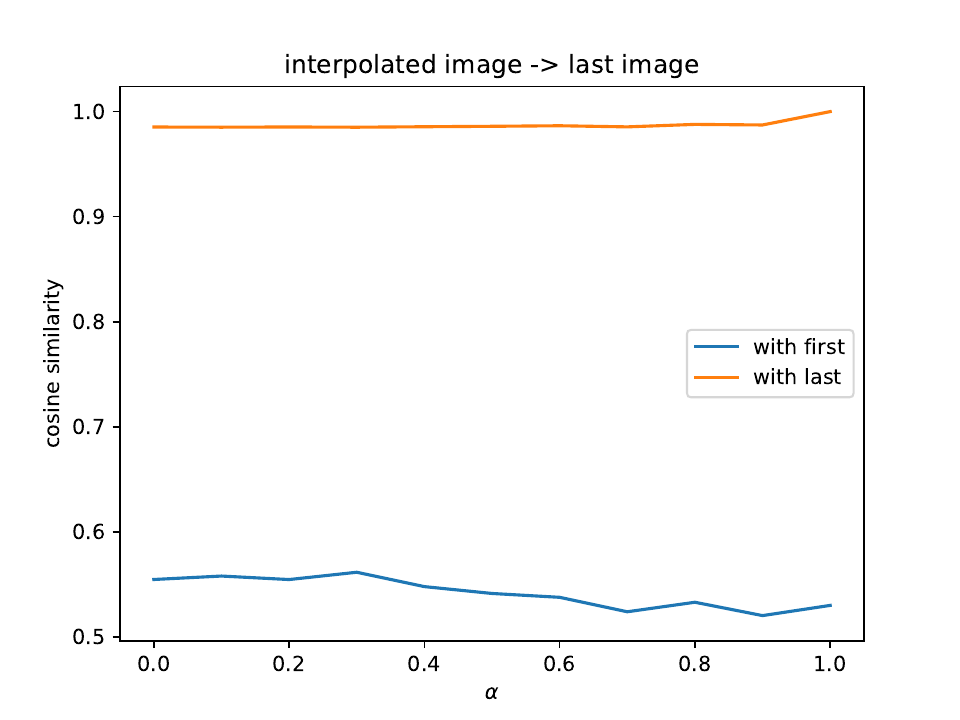}
  \vspace{-0.10in}
  \caption{(Left) The cosine similarity along the path that matches the embedding of the first image. (Right) Same as the left except along the path for that of the last image.}

  \label{fig:interpolated_similarity}
\end{figure}

\begin{figure*}[ht]
    \centering
      \includegraphics[width=0.90\textwidth]{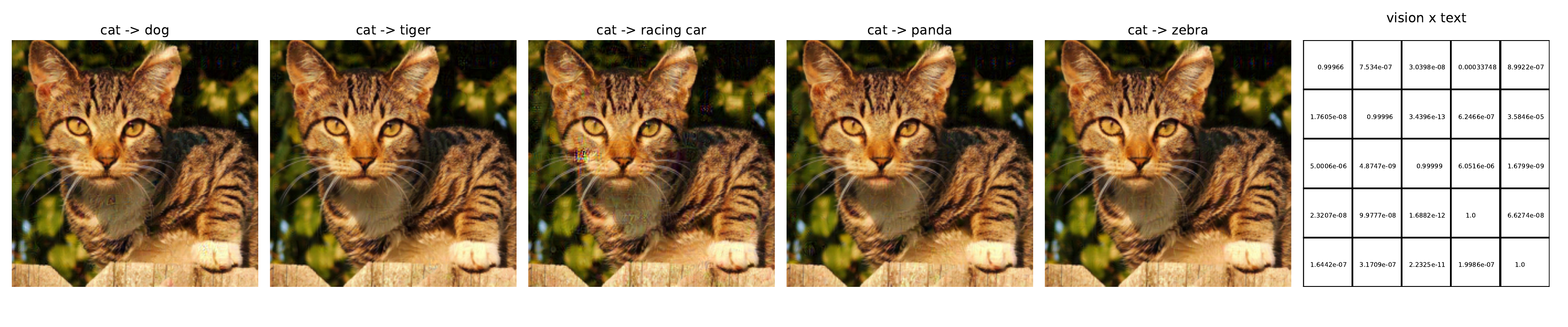}\label{fig:more1}
      \includegraphics[width=0.90\textwidth]{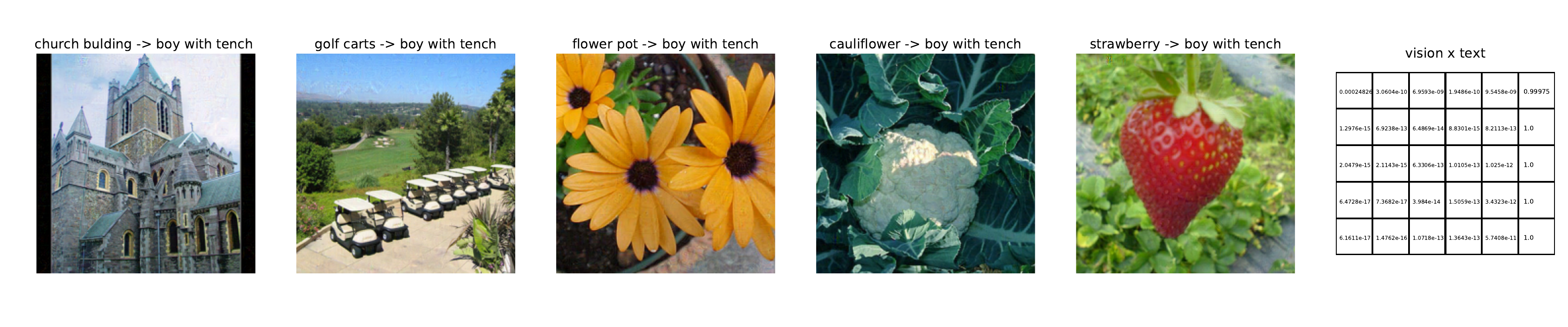}\label{fig:more2}
      \vspace{-0.20in}
    \caption{(top) More examples where visually indistinguishable images have very different embeddings and consequently are classified to other classes as in Fig. \ref{fig:overall}. Cat images are classified as a dog, a tiger, a racing car, a panda, and a zebra. (bottom) Visually very different images (e.g., a church building, some golf carts, flower pot, cauliflower, strawberry) have very similar embeddings and are classified as a boy with tench. For more details on the additional images in the first row and the second row, please see the Appendix.}
 \label{fig:more_examples}
\end{figure*}

\begin{figure}[ht]
  \centering

  \includegraphics[width=0.47\textwidth]{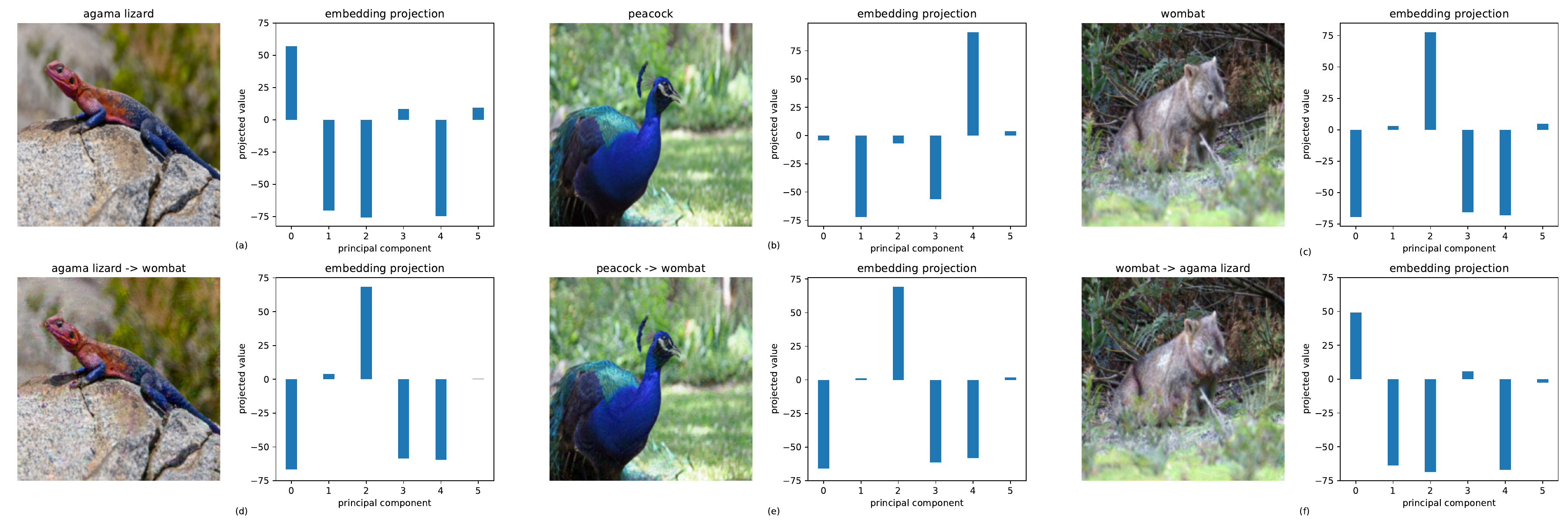}\label{fig:projection_embedding_beit}\\

  \includegraphics[width=0.47\textwidth]{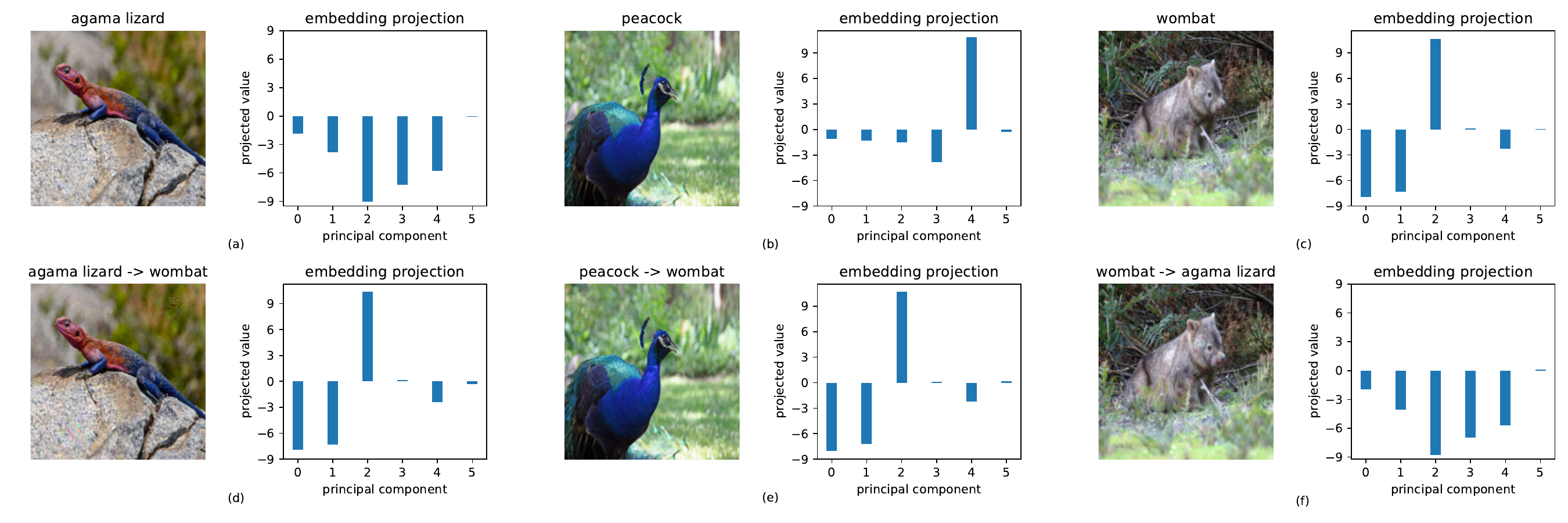}\label{fig:projection_embedding_swin}\\

  \caption{Examples obtained while the proposed framework is applied on different vision transformer models, such as (top two rows) BEiT, and (the next two rows) Swin Transformer. The results are given in the same format as depicted in Fig. \ref{fig:overall}. Additional plots for other models are also consistent and added to the Appendix. The example demonstrates that the method is model-agnostic.
}
  \label{fig:overall_diff_other_models}
  \vspace{-0.15in}
\end{figure}

As a by-product, understanding the algebraic and geometric structures of the embedding space allow us to explore the space effectively. For example, we can find adversarial attacks to the embedding of any given image using the proposed gradient procedure. Fig. \ref{fig:overall} shows three examples. To demonstrate the universal applicability of the procedure and the adversarial examples that exist almost everywhere, Fig. \ref{fig:more_examples} shows more examples from different categories from the ImageNet dataset. 
See the Appendix for additional examples. 

\textbf{Qualitative evaluation.} Our key result is that the semantic meanings of the embeddings given by transformer models are fundamentally limited as different inputs share similar embeddings while visually indistinguishable inputs have very different embeddings. As the techniques are model and dataset-agnostic, they should be effective on different transformer models and datasets, including ones for other modalities. We have conducted experiments with various vision transformer models, including MAE-like models from HuggingFace\footnotemark\footnotetext{https://huggingface.co/docs/transformers/model\_doc/beit}, such as BEiT, DEiT, Swin, ViTMAE, ViTMSN~\cite{bao2022beit,touvron2021training,liu2021swin,he2021masked,assran2022masked} and two examples are given in Fig.~\ref{fig:overall_diff_other_models}. Please refer to the Appendix for additional results with other models and datasets. 


In general, as shown in the examples, the proposed technique works well with any randomly chosen image from a different target class. Additionally, Fig. 3 displays the singular values of the Jacobian matrix, revealing notable differences in the singular value distributions between the original and manipulated images.


So far, we have shown the structures of the embedding space of a particular point using concrete examples. Our framework allows us to explore the paths and the space more broadly. 
Fig. \ref{fig:nonlinearity} shows how the embeddings change as the input changes from one image to the one that matches a specified target but remains visually indistinguishable. The plot shows the changes roughly linearly.

Compared to existing adversarial attack methods, one distinctive feature of our proposed framework is that we can exploit how different subspaces are connected. Fig. \ref{fig:interpolated_similarity} shows one such path example. By applying the match-finding procedure, we are able to construct and connect different subspaces. 

The results show that the embeddings are inherently limited semantically when analyzed systematically. 
Locally, the model is sensitive to small changes along the directions in the normal space. In the null space, the embeddings remain constant while the input changes substantially. By connecting local null spaces, we have connected spaces where embeddings are similar, but inputs can be very different. As the embedding space is high dimensional in nature, testing using datasets is inherently limited. The systematic analyses are essential.

\section{Discussion}
By using computational procedures with mathematical analyses, we characterize the embedding space of a vision transformer both locally and globally. Note that the proposed framework can be applied to characterizing any model directly as long as the input varies continuously so that the Jacobian can be estimated properly. With multimodal models, the framework can also be used to study other models with discrete inputs indirectly via other joint embeddings. 

It may be attempting to categorize our framework as an adversarial attack technique. Our primary focus is on analyzing the embedding space; we utilize the ImageBind solely as a classifier to validate our findings and is not used otherwise. While our embedding matching procedure can be used to generate effective adversarial examples, it is fundamentally different. Our technique is classifier agnostic and does not exploit features specific to classifiers. Consequently, our examples with matched embeddings will appear to be the same to any classifier or downstream model that builds on embeddings. On the other hand, traditional adversarial attacks are specific to classifiers and applications, focusing on altering their outputs by changing the input.

The plausible root cause of such adversarial examples and also semantically different images with identical embeddings is that transformers do not require the inputs to be aligned to have similar embeddings. By adding alignment-sensitive components to the embedding could mitigate the problem, which is being investigated further. Additionally, based on the singular values of the Jacobian matrix, it appears possible to evaluate the robustness of the models, which is being investigated. 

The results shown in this paper seem not to be consistent with the impressive results demonstrated by such models. Note that almost all existing results are measured on benchmark datasets. Due to the high dimensionality of the embedding space and the input space, even the largest dataset will cover the spaces very sparsely. We believe that systematic evaluations such as ours are necessary if one likes to evaluate models to be able to predict their behaviors in the entire space rather than on samples. 

Note that the problem of how to estimate the global and local Lipschitz constants of neural networks and transformers has been studied mathematically. In particular, LipsFormer \cite{qi2023lipsformer} shows that degenerated cases can cause the Lipschitz constant to be unbounded. However, none of these techniques have been scaled to the large models that are being deployed, including the ones we have used. Our results are also complementary in nature; we show the distributions of the local directional Lipschitz constants of real trained large models and are able to estimate them accurately using the Jacobian matrix. For applications, the Lipschitz constants themselves provide an upper bound of the rate of the change and may not be sufficient to understand their behavior for typical inputs.

\section{Conclusion}
In this paper, we show the structures of the embedding spaces using algorithms and mathematical analyses. It is attempting to conclude that recent pre-trained models can be used to build any effective applications based on their performance on benchmark datasets. While such models give impressive performance, their inherent generalization abilities are limited by the properties of the underlying embedding spaces. Before this fundamental limitation can be addressed, such models should not be used for critical applications.

{\small
\bibliography{reference2}
}

\clearpage

\appendix

\section{Appendix}

\subsection{More on Vision Transformers}

Very recently, several multi-modal models have been introduced~\cite{xu2023multimodal,zhu2023minigpt,Gpt42023Open,girdhar2023imagebind}. By using a shared embedding space among different modalities, such joint models have shown to have advantages. Vision transformers have been successful in various vision tasks due to their ability to treat an image as a sequence of patches and utilize self-attention mechanisms.

\begin{figure}[H]
  \centering
    \includegraphics[width=0.90\columnwidth]{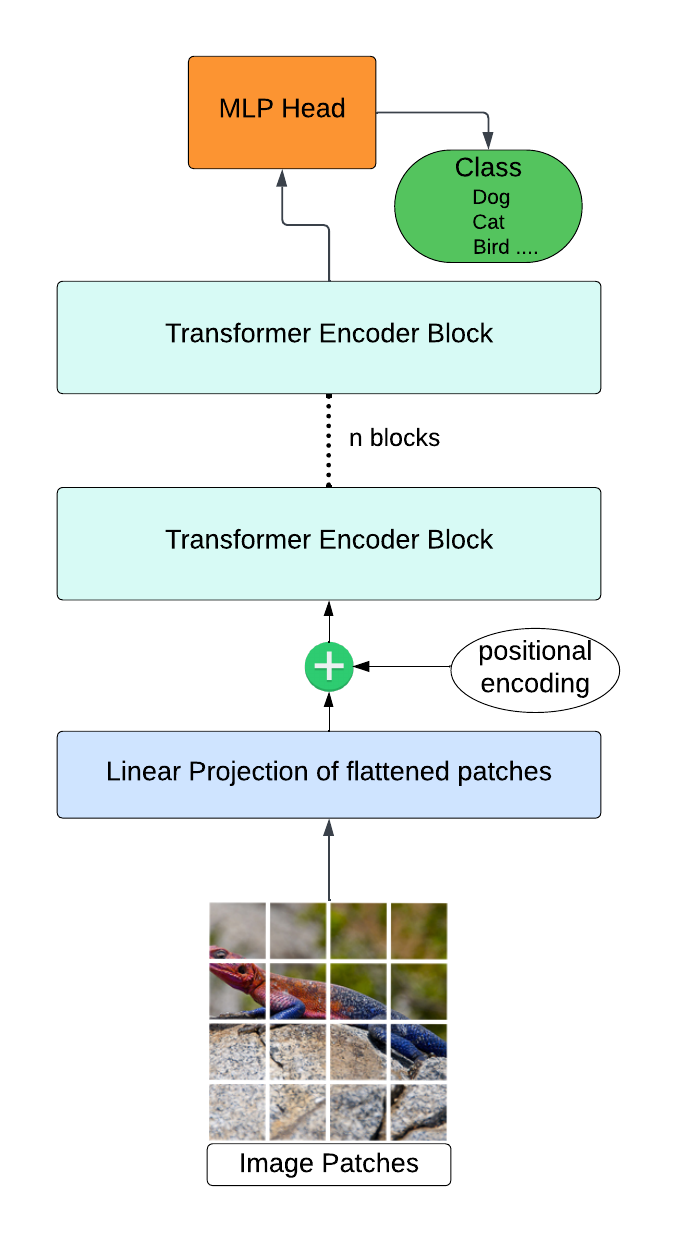}
  \caption{Vision Transformer (ViT) architecture~\cite{dosovitskiy2021image}.}
  \label{fig:vt_arch}
\end{figure}

A collection of transformer blocks make up the Vision Transformer Architecture. Each transformer block comprises two sub-layers: a multi-headed self-attention layer and a feed-forward layer. The self-attention layer computes attention weights for each pixel in the image based on its relationship with all other pixels, while the feed-forward layer applies a non-linear transformation to the self-attention layer's output. The patch embedding layer separates the image into fixed-size patches before mapping each patch to a high-dimensional vector representation. These patch embeddings are then supplied into the transformer blocks to be processed further \cite{dosovitskiy2021image}.

\subsection{Additional Results}
Here we provide more details and additional information about the results we have included in the main text.

\begin{figure*}[ht]
  \centering
  \includegraphics[width=0.99\textwidth]{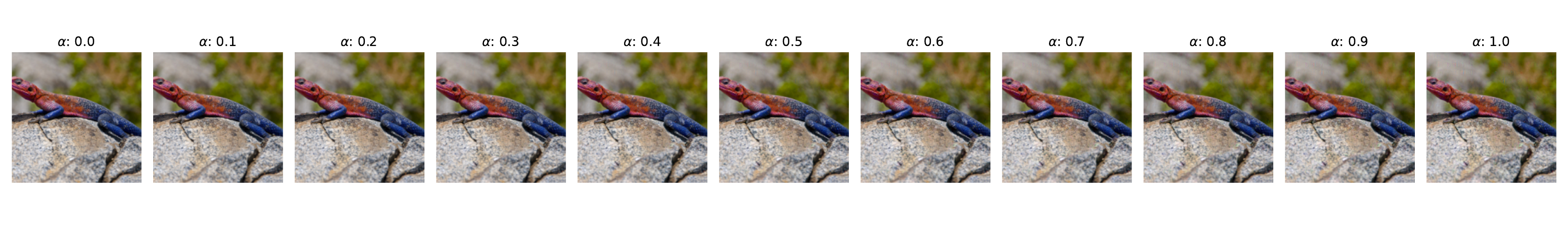}\label{fig:for_7}\\

  \caption{Interpolated Images for Fig. \ref{fig:nonlinearity}. 
}
  \label{fig:for_fig_7}
\end{figure*}

\begin{figure*}[ht]
  \centering
  \includegraphics[width=0.99\textwidth]{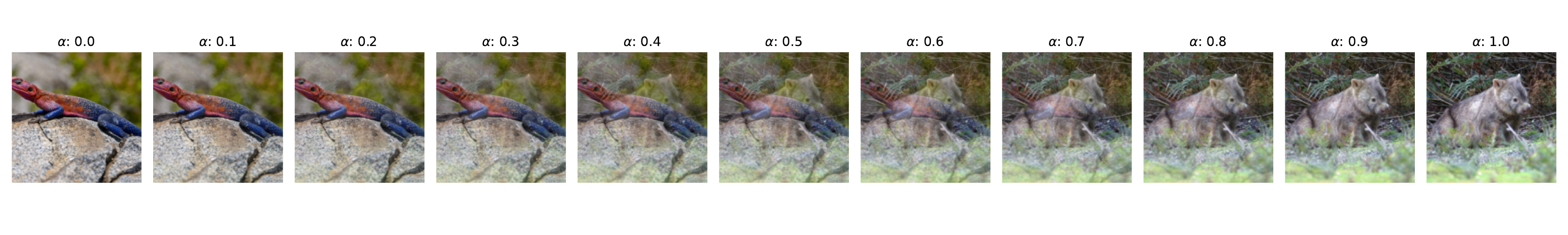}\label{fig:for_8}\\

  \caption{Interpolated Images for Fig. \ref{fig:interpolated_similarity}. 
}
  \label{fig:for_fig_8}
\end{figure*}

Fig. \ref{fig:for_fig_7} and Fig. \ref{fig:for_fig_8} show the interpolated images along the path. 

\begin{figure*}[ht]
  \centering

  \includegraphics[width=0.90\textwidth]
  {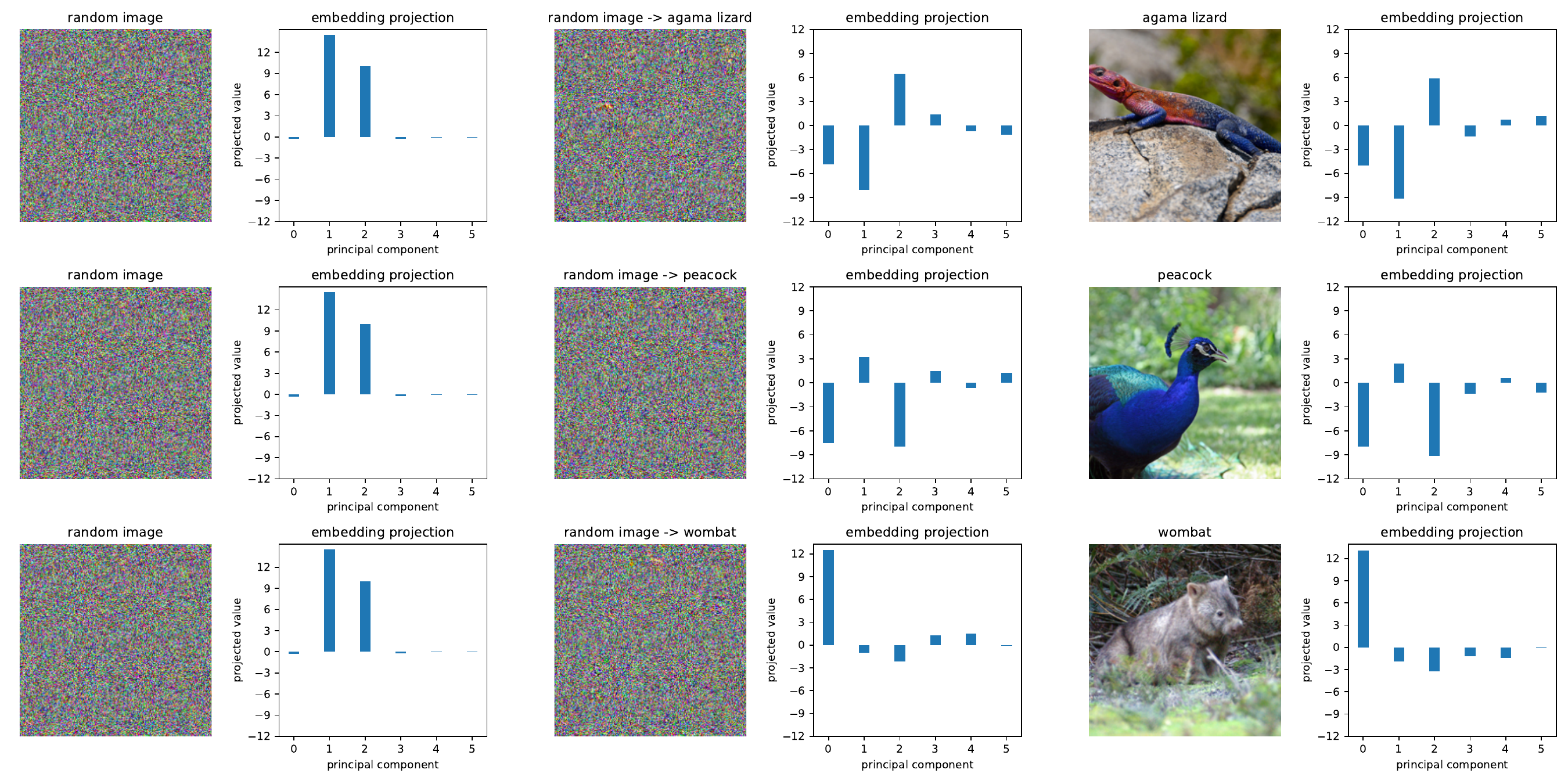}

  \caption{Example of random image (left) that matches a target embedding (right), 
  with the final image shown in the middle.}
  \label{fig:randtoreal}
\end{figure*}

To further demonstrate the effectiveness of the gradient procedure to match embeddings, we have applied the procedure to numerous images from different sources. As random images are typical in the input image space, we have applied the procedure to match a specified embedding from randomly generated images. Fig. \ref{fig:randtoreal} shows that we can match the embeddings of images from a random image; These results, along with outcomes from other datasets, demonstrate the efficacy of our technique across all the images we have utilized.

\begin{figure*}[ht]
  \centering
  \subfloat[]
  {\includegraphics[width=0.90\textwidth]{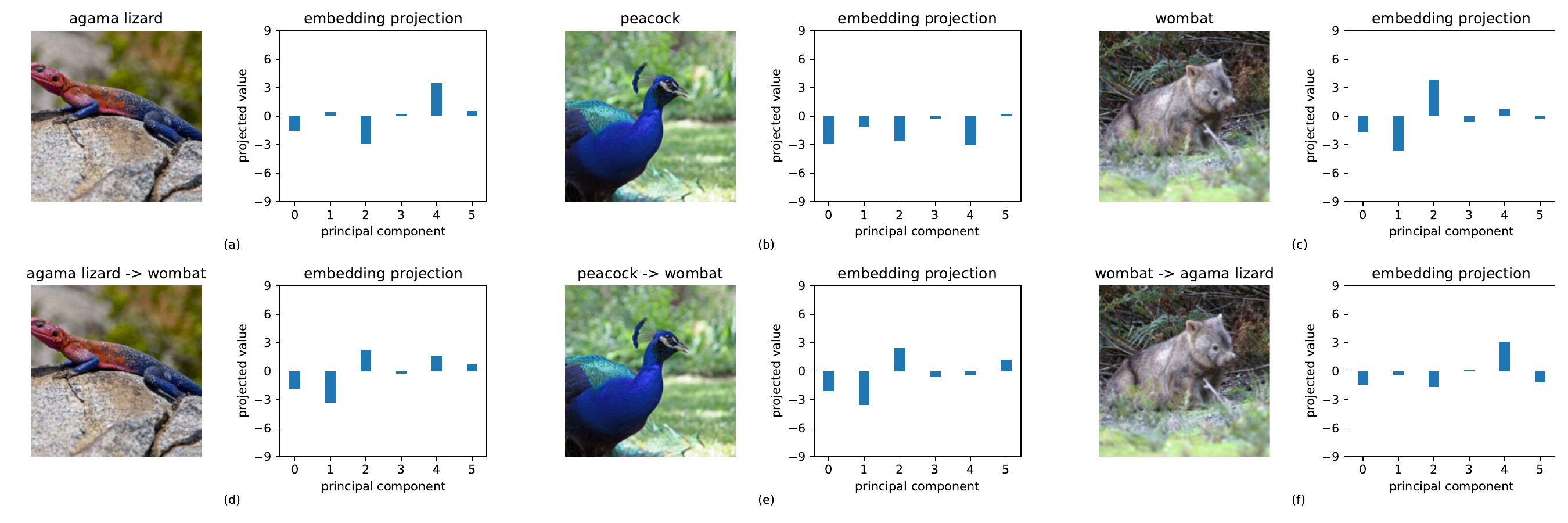}\label{fig:projection_embedding_b16}}\\

  \subfloat[]
  {\includegraphics[width=0.90\textwidth]{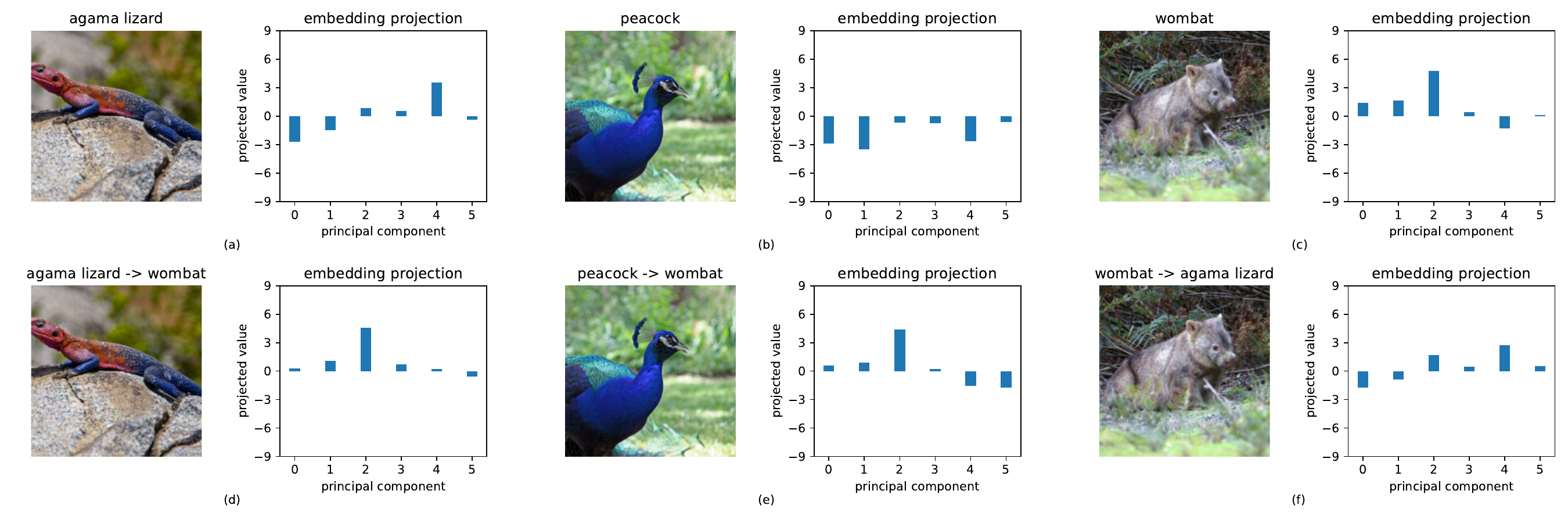}\label{fig:projection_embedding_b32}}\\

  \subfloat[]
  {\includegraphics[width=0.90\textwidth]{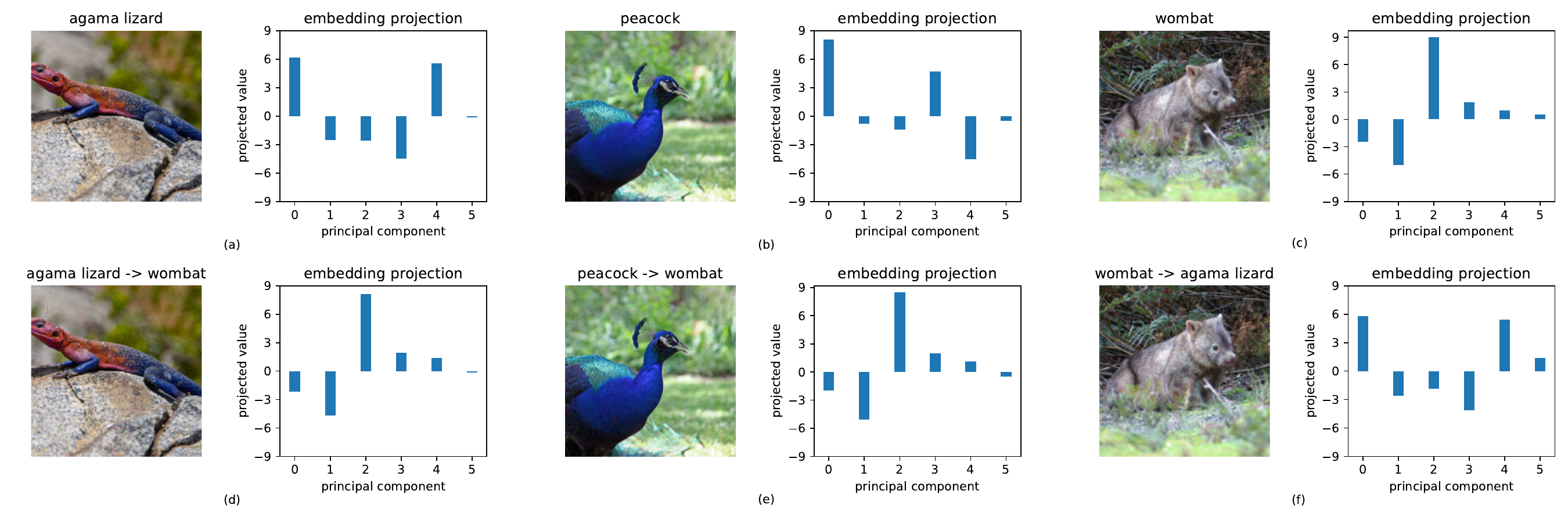}\label{fig:projection_embedding_l14}}

  \caption{More examples from ImageNet obtained using the proposed framework with different variants of the original vision transformer, such as (top) ViT-B-16, which has the embedding dimension of 512, (center) ViT-B-32, which has the embedding dimension of 512, (bottom) ViT-L-14 which has the embedding dimension of 768. 
}
  \label{fig:overall_diff_vits}
\end{figure*}

\begin{figure*}[ht]
  \centering

  \includegraphics[width=0.90\textwidth]{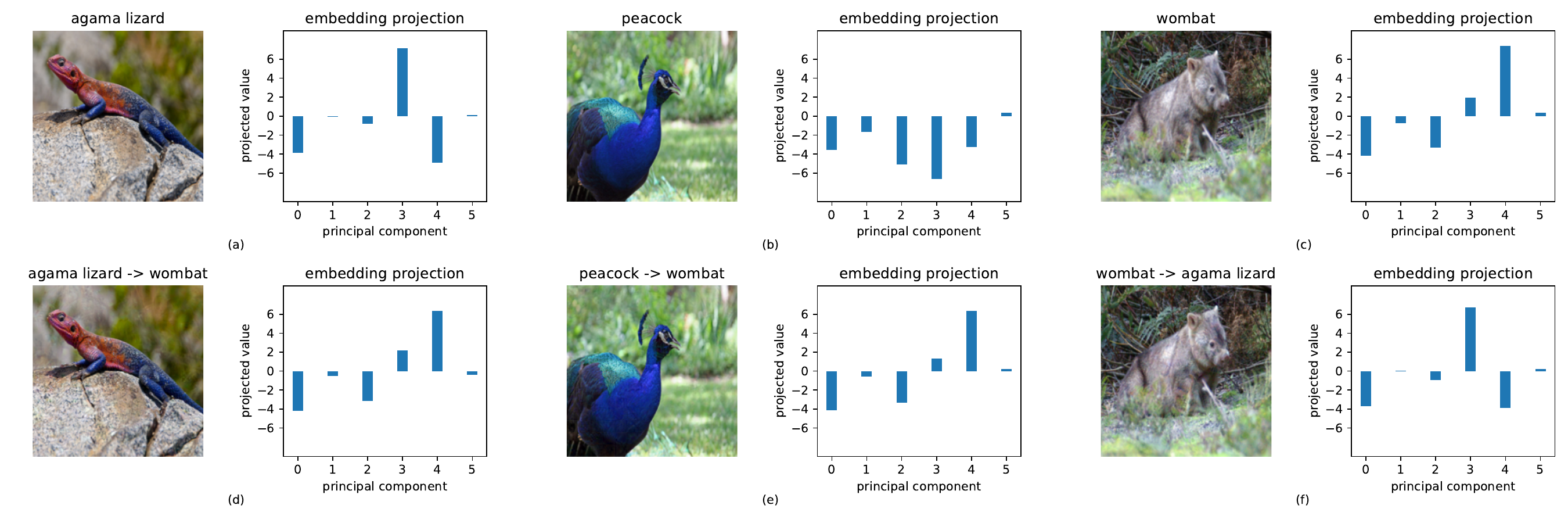}\label{fig:projection_embedding_deit}\\

  \includegraphics[width=0.90\textwidth]{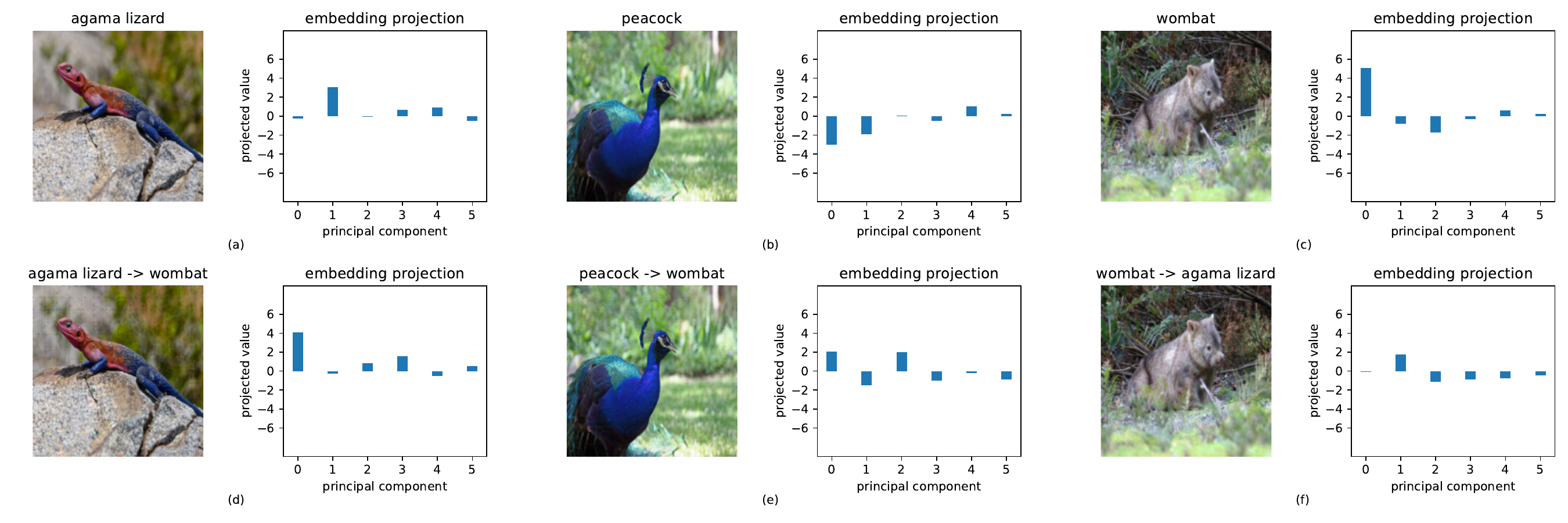}\label{fig:projection_embedding_vitmae}\\

  \includegraphics[width=0.90\textwidth]{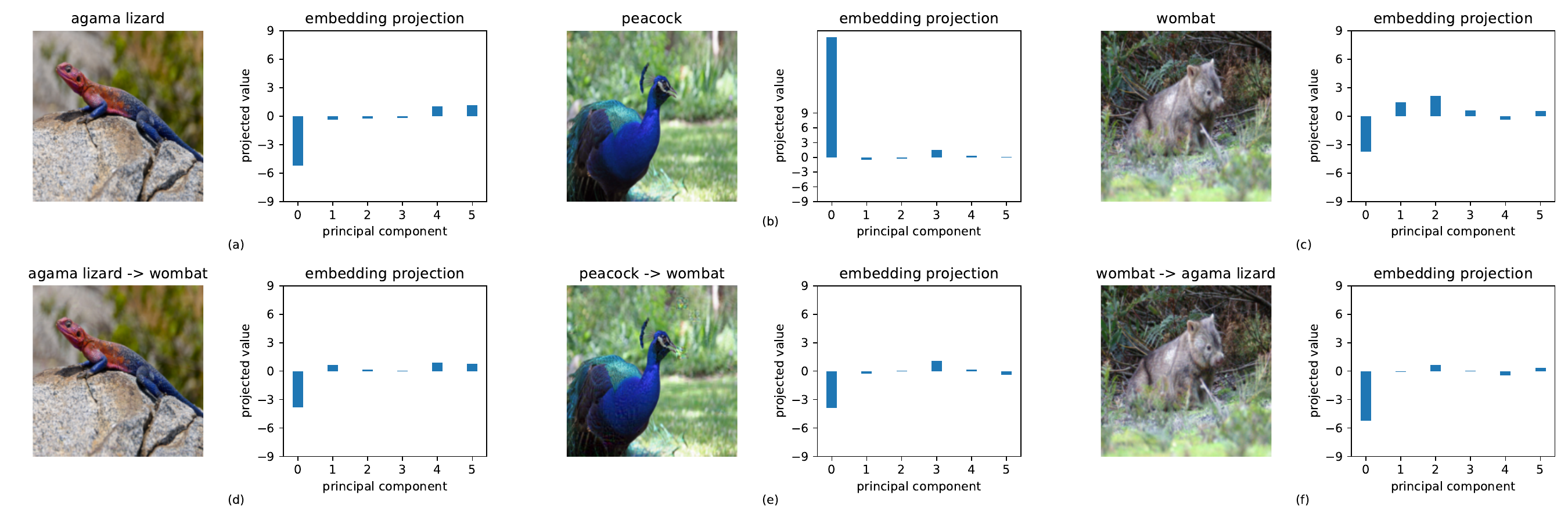}\label{fig:projection_embedding_vitmsn}

  \caption{Same as Fig. \ref{fig:overall} and Fig. \ref{fig:overall_diff_other_models}, in support of demonstrating that the proposed framework is model-agnostic; shown for different other vision transformer models, such as (top two rows) DEiT, (middle two rows) ViTMAE and (the next two rows) ViTMSN.
}
  \label{fig:overall_diff_other_models_more}
  \vspace{-0.15in}
\end{figure*}

In the main paper, the results are generated using the pre-trained ImageBind~\cite{girdhar2023imagebind} model, which utilizes a pre-trained CLIP model (ViT-H-14). As the framework does not rely on the specifics of the ImageBind, it is effective for other models and datasets as well. To demonstrate that our framework works equally well with other variants, 
Fig. \ref{fig:overall_diff_vits} shows the results on several different variants of the original vision transformer models\footnotemark\footnotetext{https://github.com/openai/CLIP}. To further showcase the model-agnostic nature of our techniques, we conduct experiments with diverse vision transformer models, including DEiT, ViTMAE, and ViTMSN. Please refer to Fig. \ref{fig:overall_diff_other_models} in the main paper and Fig. \ref{fig:overall_diff_other_models_more} for detailed results. 

\begin{figure*}[ht]
    \centering
      \includegraphics[width=0.950\textwidth]{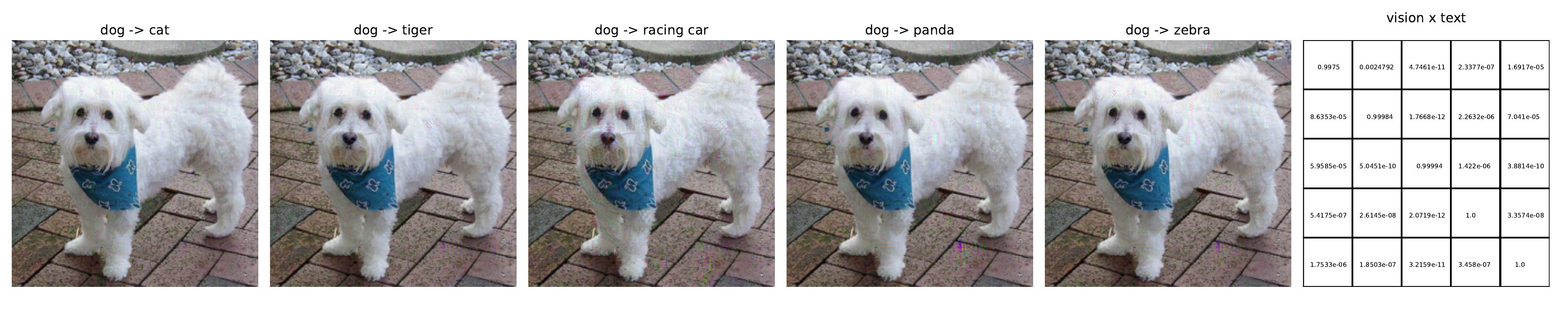}\label{fig:more3}
      \includegraphics[width=0.950\textwidth]{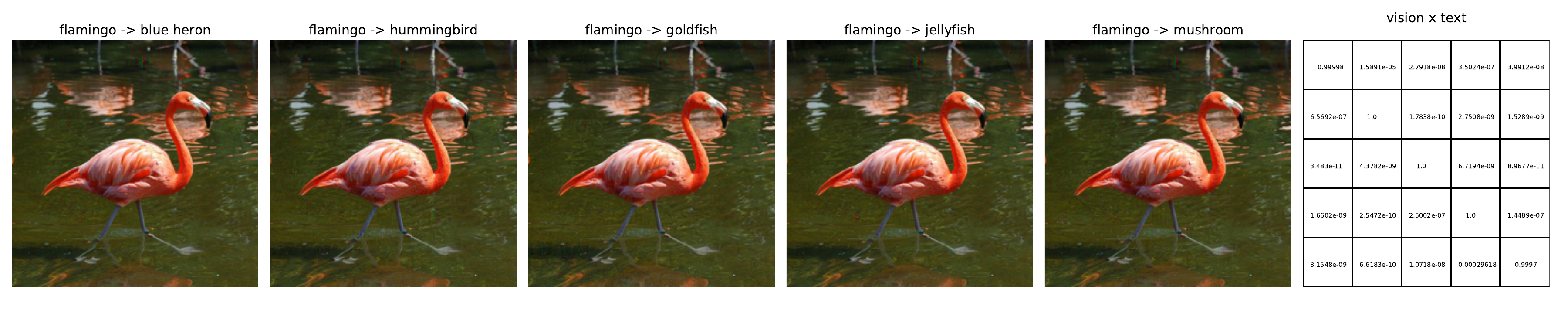}\label{fig:more4}
      \includegraphics[width=0.950\textwidth]{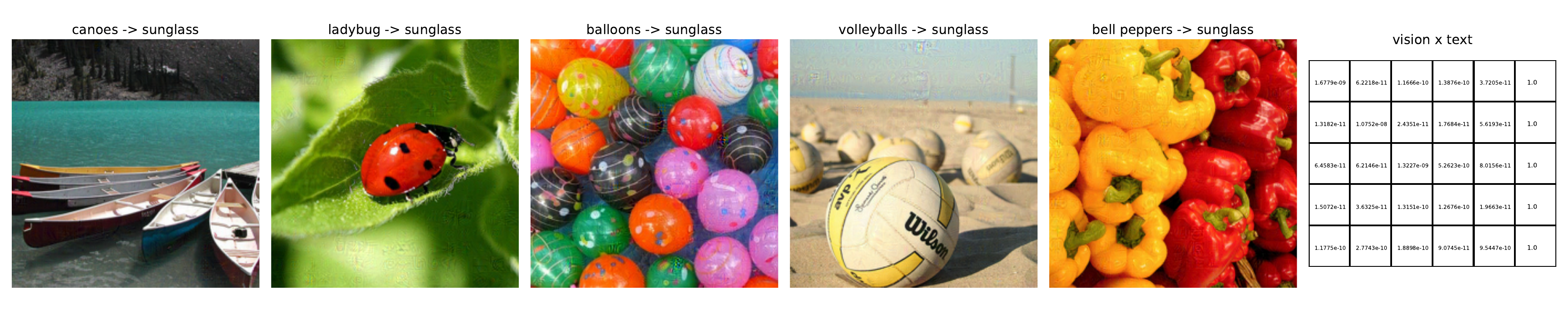}\label{fig:more5}
      \includegraphics[width=0.950\textwidth]{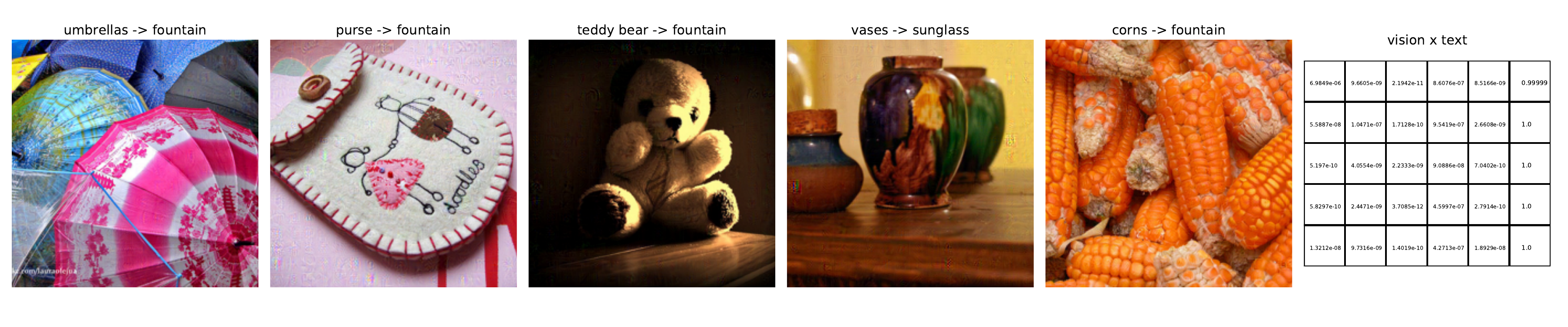}\label{fig:more6}

    \caption{(first row) Additional examples where visually indistinguishable images have very different embeddings and consequently are classified to other classes as in Fig. \ref{fig:overall} and Fig. \ref{fig:more_examples}. Dog images are classified as a cat, a tiger, a racing car, a panda, and a zebra. (second row) Similar as first row, flamingo images are classified as a heron, hummingbird, goldfish, jellyfish, and mushroom. (third row) Visually very different images (e.g., some canoes, a ladybug, some balloons, some volleyballs, some bell peppers) have very similar embeddings and are classified as sunglass. (fourth row) Similar as third row, different images (e.g., some umbrellas, a purse, a teddy bear, some vases, some corns) are classified as fountain. The examples are strictly randomly chosen. There is no postselection involved.}
 \label{fig:more_examples_2}
\end{figure*}

\begin{figure*}[ht]
    \centering
      \includegraphics[width=0.950\textwidth]{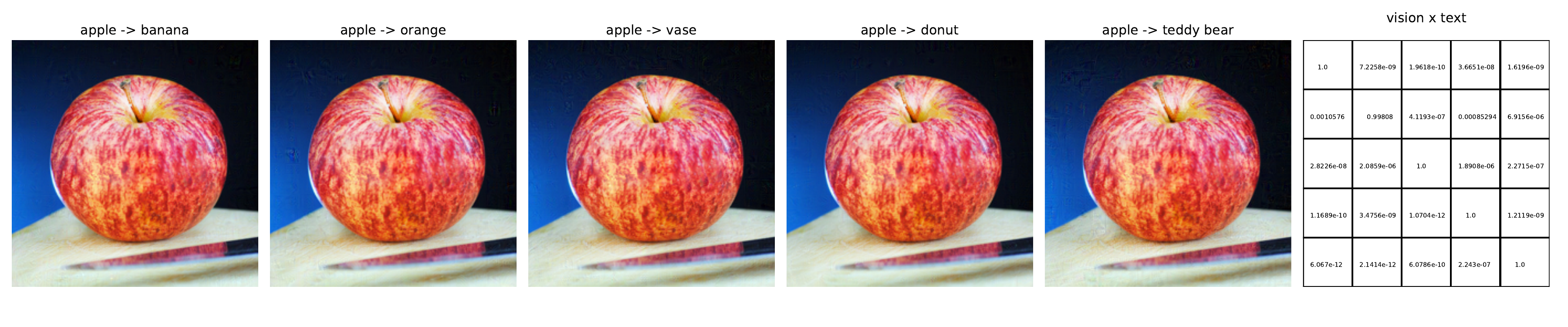}\label{fig:more7}
  
      \includegraphics[width=0.950\textwidth]{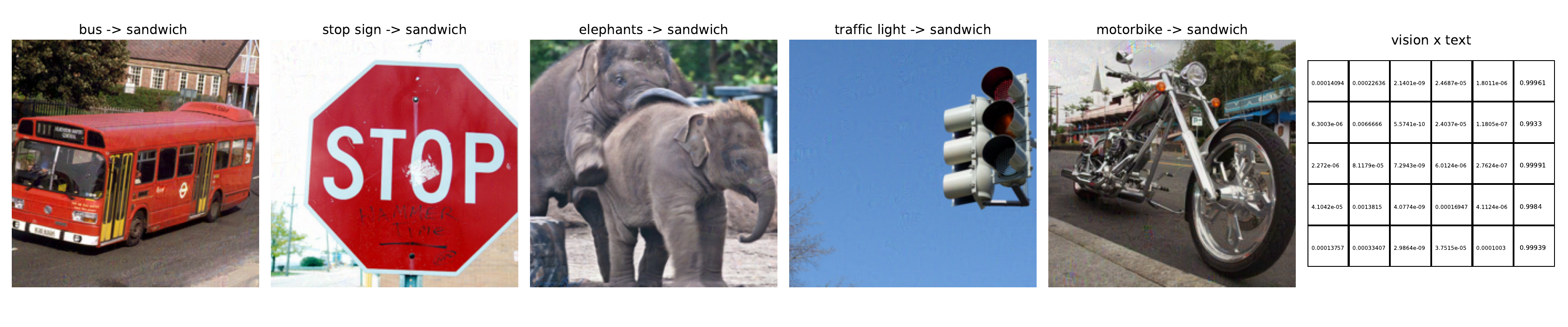}\label{fig:more8}

    \caption{More examples involving MS-COCO dataset. (top) Visually indistinguishable images have very different representations via embedding alignment with the corresponding images and therefore very different classification outcomes. (bottom) Visually very different images have very similar embeddings, aligned to the embedding of a specific image and classified into the corresponding class. Again the samples are randomly chosen.}
 \label{fig:more_examples_3}
\end{figure*}

\begin{figure*}[ht]
    \centering
      \includegraphics[width=0.950\textwidth]{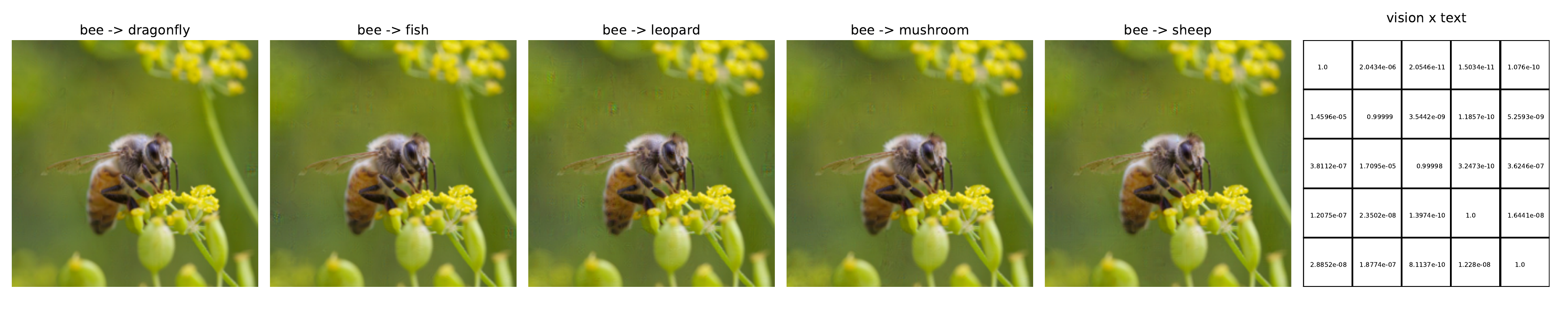}\label{fig:more9}
  
      \includegraphics[width=0.950\textwidth]{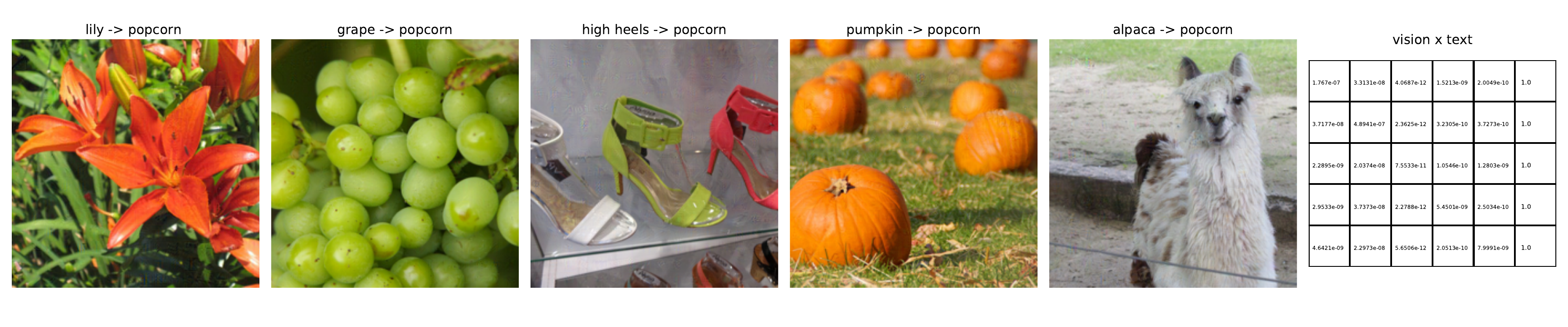}\label{fig:more10}

    \caption{More examples involving Open Images dataset having high-resolution images. (top) Visually indistinguishable images have very different representations via embedding alignment with the corresponding images and therefore very different classification outcomes. (bottom) Visually very different images have very similar embeddings, aligned to the embedding of a specific image and classified into the corresponding class. The samples are randomly chosen.}
 \label{fig:more_examples_4}
\end{figure*}

Fig. \ref{fig:more_examples_2} provides more examples on ImageNet, where visually indistinguishable images have very different embeddings and consequently are classified into other classes. In contrast, visually very different images have very similar embeddings, aligned to the embedding of a particular image and classified into the corresponding class. Additionally, in Fig. \ref{fig:more_examples_3} and Fig. \ref{fig:more_examples_4}, we present further examples applying our proposed framework to the MS-COCO and Open Images datasets, affirming the dataset-agnostic nature of our approach.

\begin{figure*}[ht]
  \centering

  \includegraphics[width=0.14\textwidth]
  {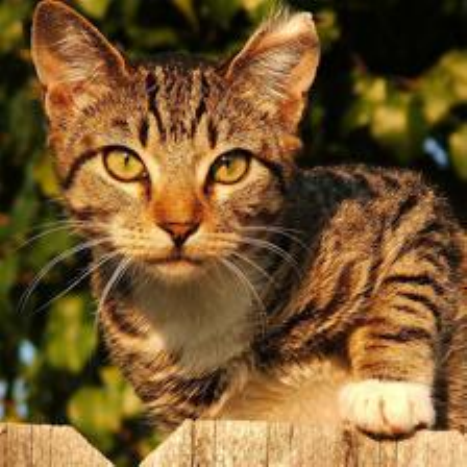}
  \includegraphics[width=0.14\textwidth]
  {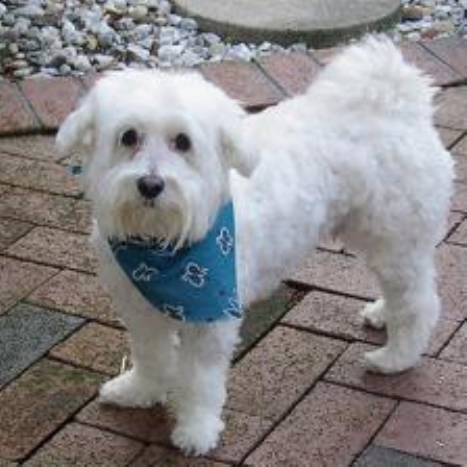}
  \includegraphics[width=0.14\textwidth]
  {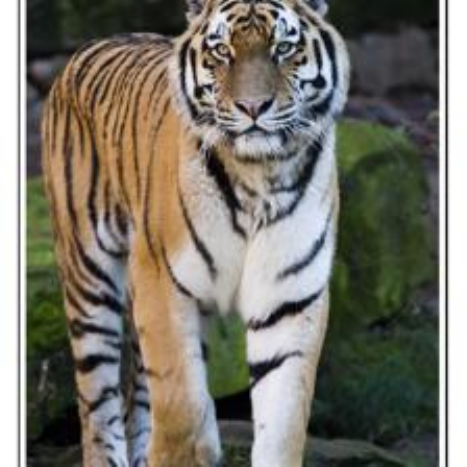}
  \includegraphics[width=0.14\textwidth]
  {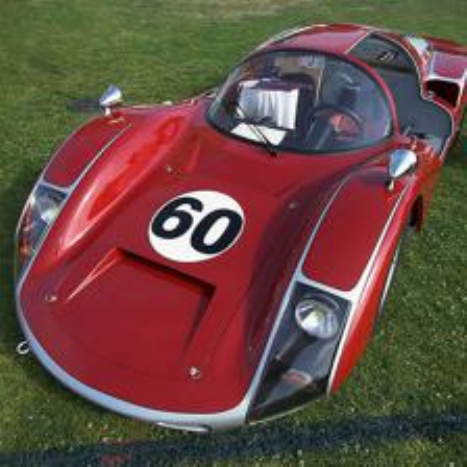}
  \includegraphics[width=0.14\textwidth]
  {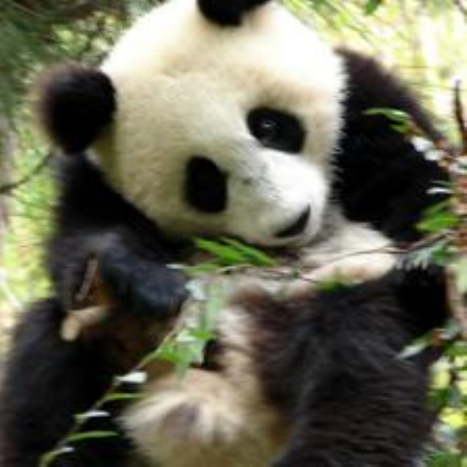}
  \includegraphics[width=0.14\textwidth]
  {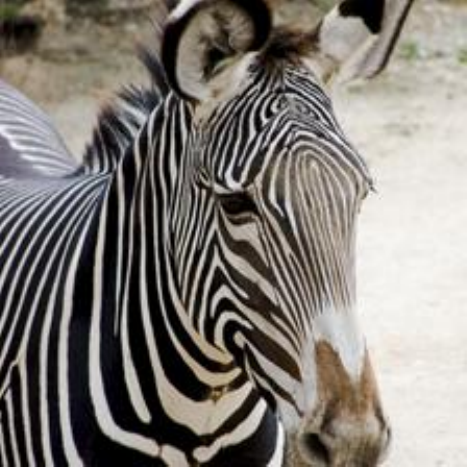}\\

  \includegraphics[width=0.14\textwidth]
  {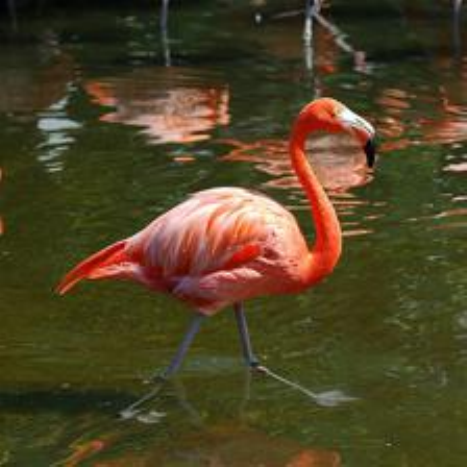}
  \includegraphics[width=0.14\textwidth]
  {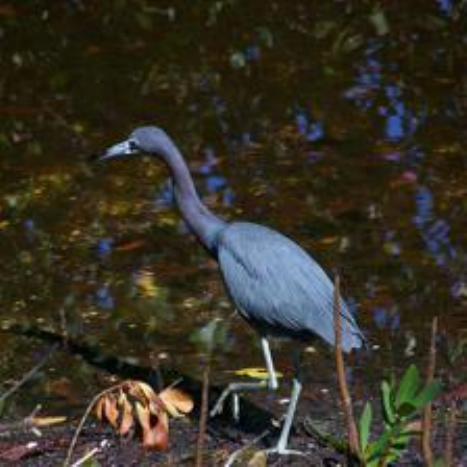}
  \includegraphics[width=0.14\textwidth]
  {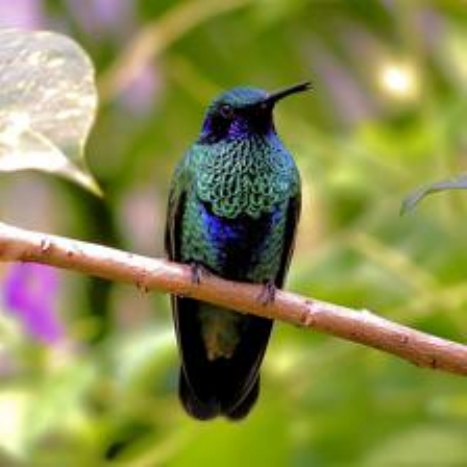}
  \includegraphics[width=0.14\textwidth]
  {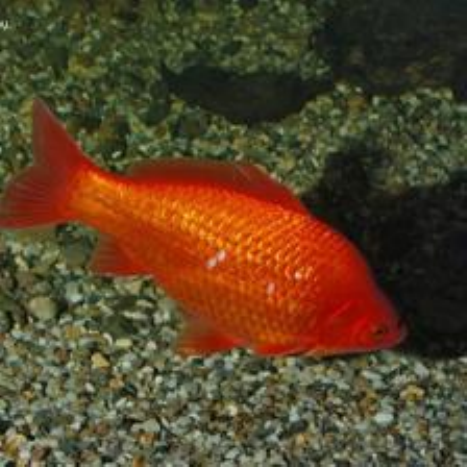}
  \includegraphics[width=0.14\textwidth]
  {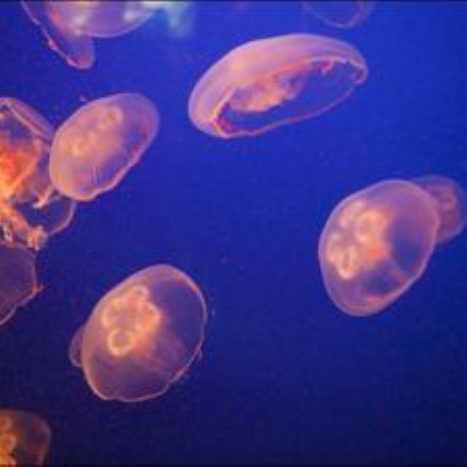}
  \includegraphics[width=0.14\textwidth]
  {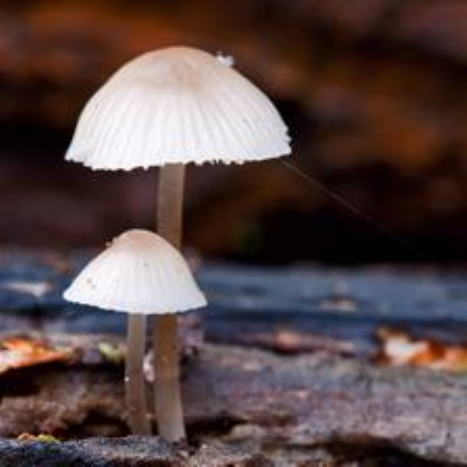}\\

  \includegraphics[width=0.14\textwidth]
  {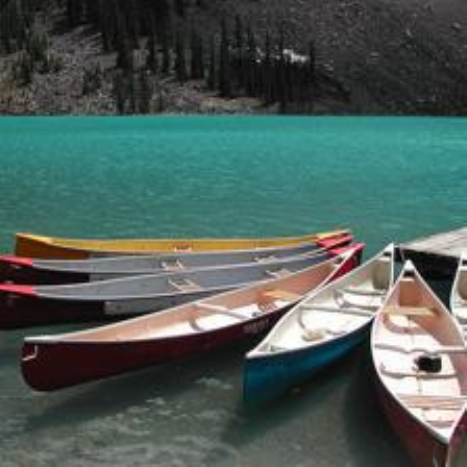}
  \includegraphics[width=0.14\textwidth]
  {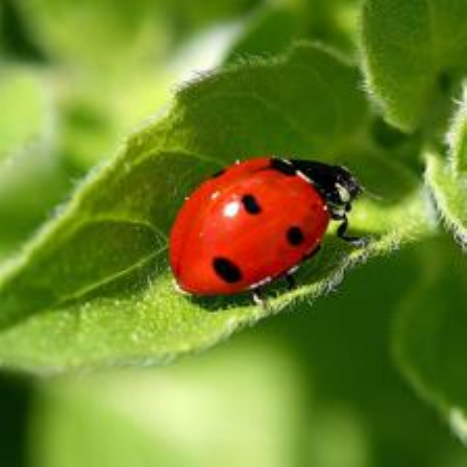}
  \includegraphics[width=0.14\textwidth]
  {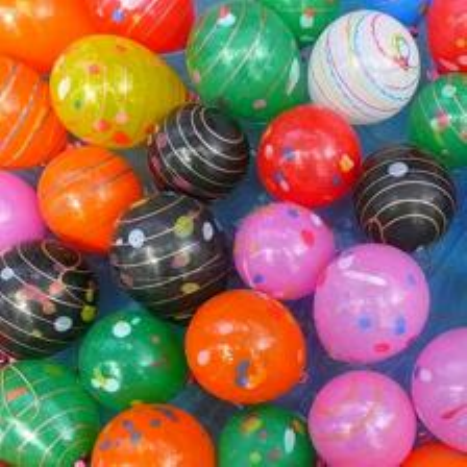}
  \includegraphics[width=0.14\textwidth]
  {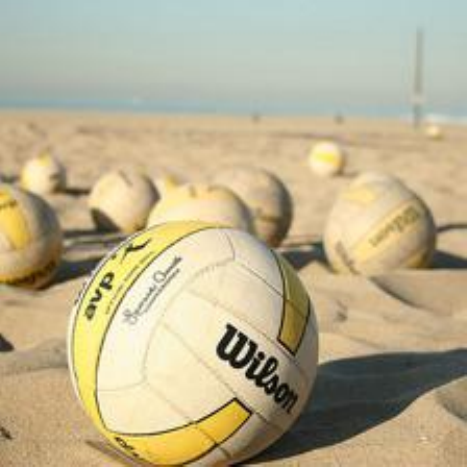}
  \includegraphics[width=0.14\textwidth]
  {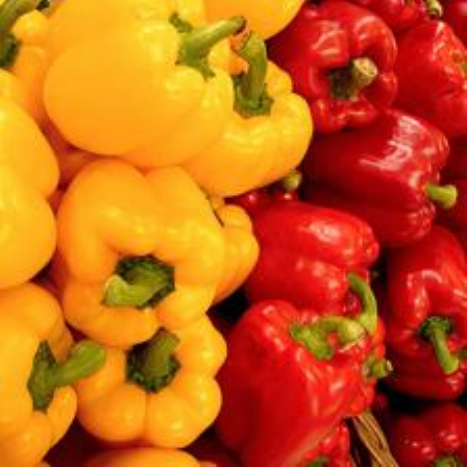}
  \includegraphics[width=0.14\textwidth]
  {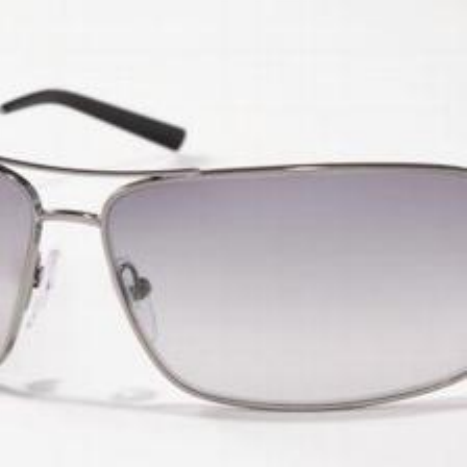}\\

  \includegraphics[width=0.14\textwidth]
  {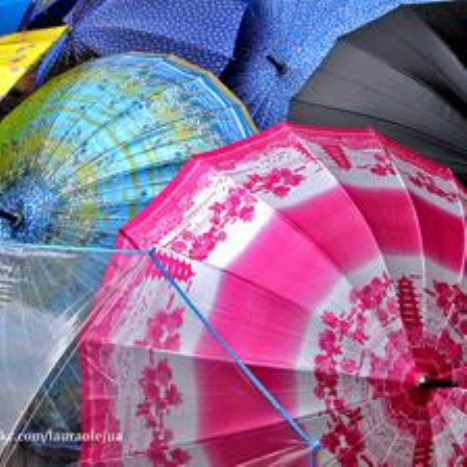}
  \includegraphics[width=0.14\textwidth]
  {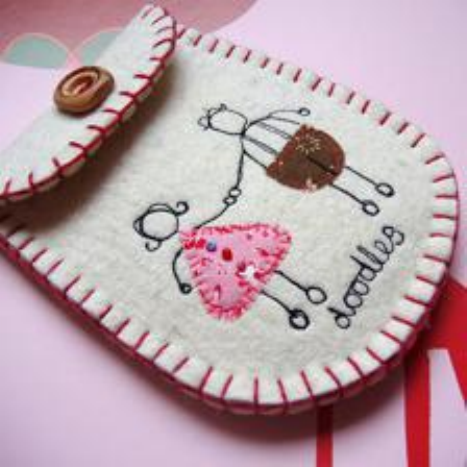}
  \includegraphics[width=0.14\textwidth]
  {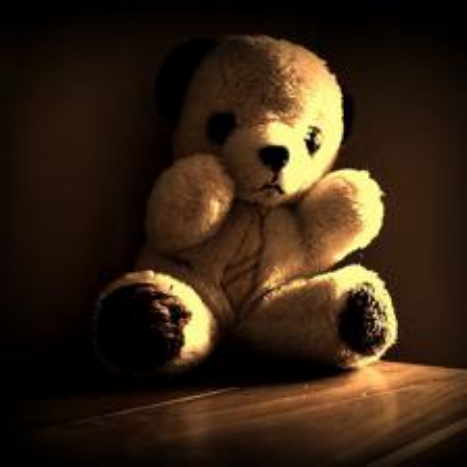}
  \includegraphics[width=0.14\textwidth]
  {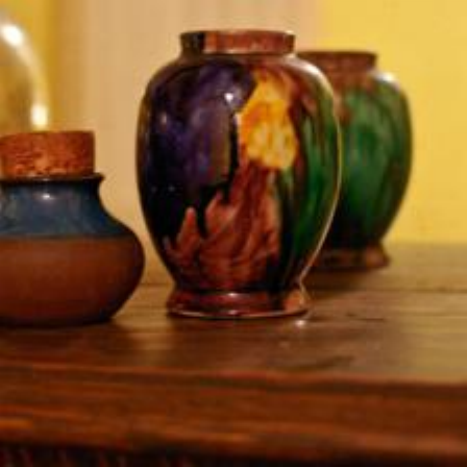}
  \includegraphics[width=0.14\textwidth]
  {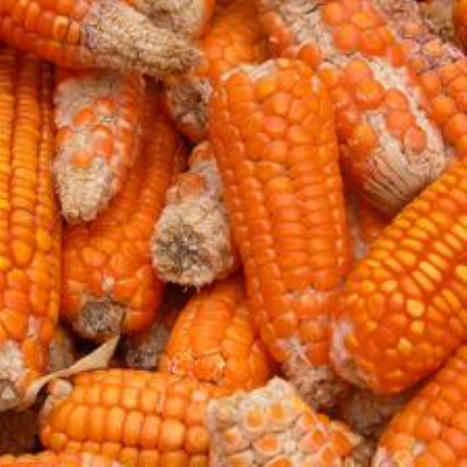}
  \includegraphics[width=0.14\textwidth]
  {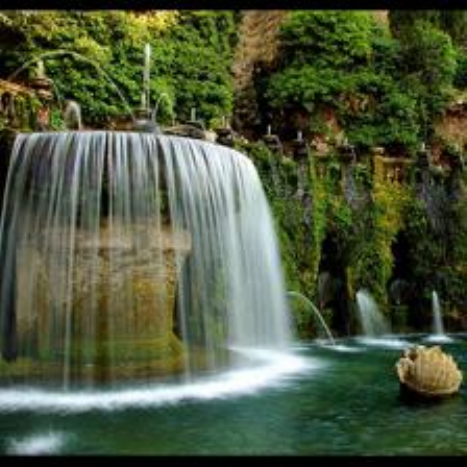}

  \caption{The original images from ImageNet corresponds to Fig. \ref{fig:more_examples} and Fig \ref{fig:more_examples_2}.}
  \label{fig:originals_imagenet}
\end{figure*}

\begin{figure*}[ht]
  \centering

  \includegraphics[width=0.14\textwidth]
  {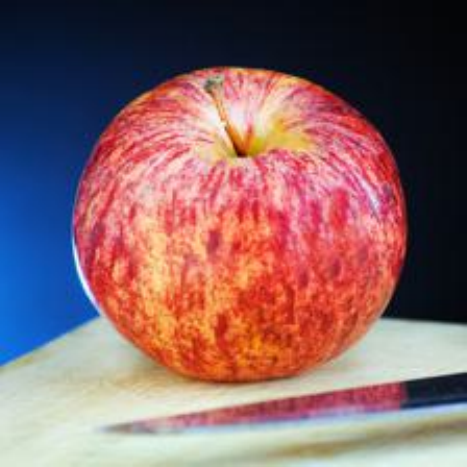}
  \includegraphics[width=0.14\textwidth]
  {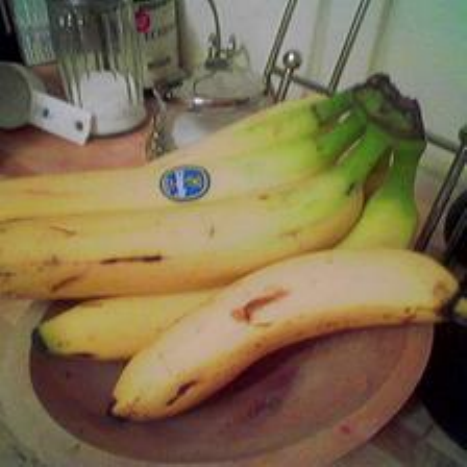}
  \includegraphics[width=0.14\textwidth]
  {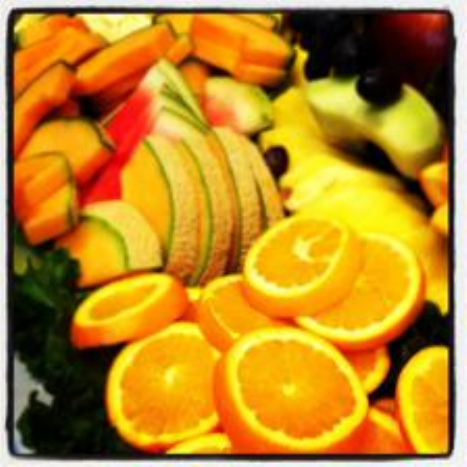}
  \includegraphics[width=0.14\textwidth]
  {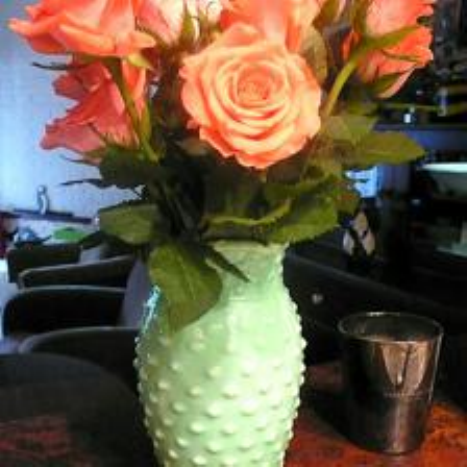}
  \includegraphics[width=0.14\textwidth]
  {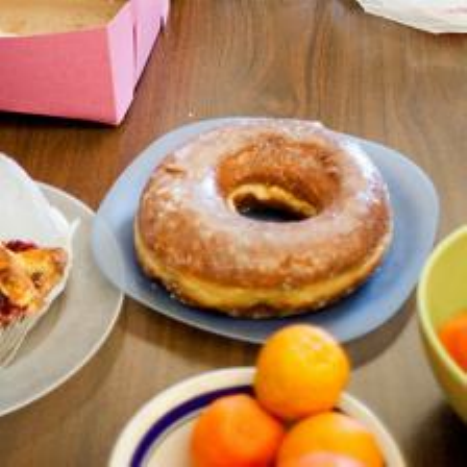}
  \includegraphics[width=0.14\textwidth]
  {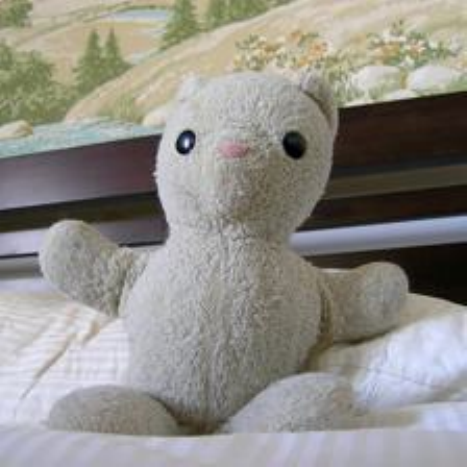}\\

  \includegraphics[width=0.14\textwidth]
  {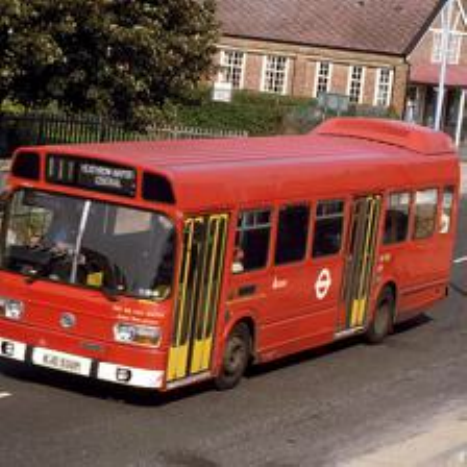}
  \includegraphics[width=0.14\textwidth]
  {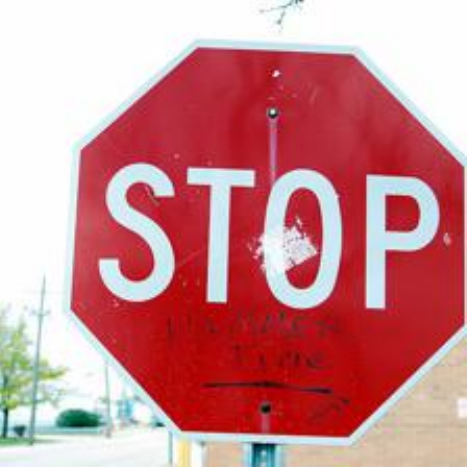}
  \includegraphics[width=0.14\textwidth]
  {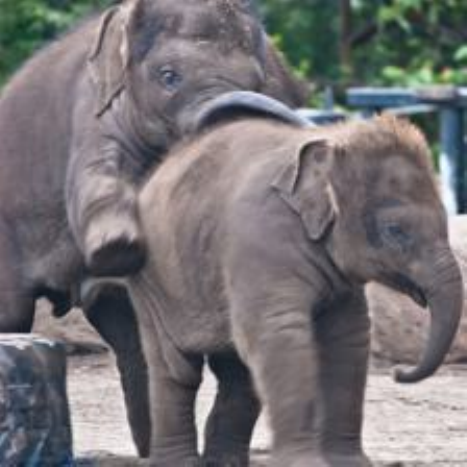}
  \includegraphics[width=0.14\textwidth]
  {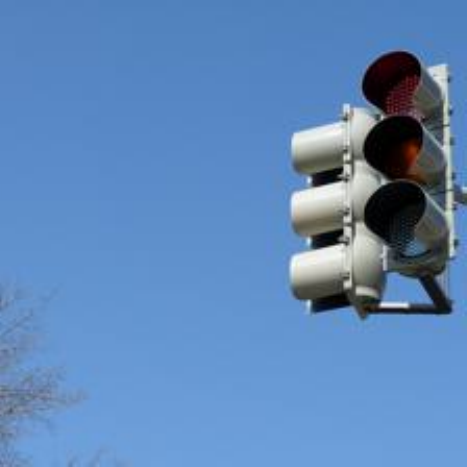}
  \includegraphics[width=0.14\textwidth]
  {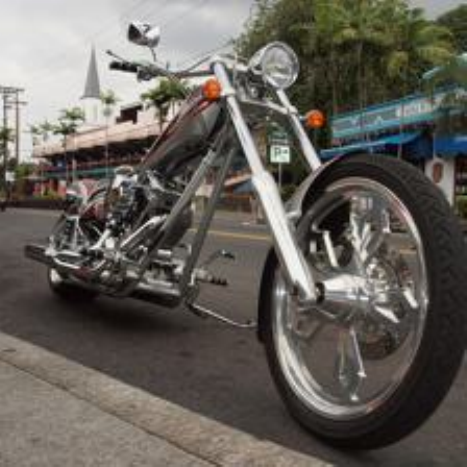}
  \includegraphics[width=0.14\textwidth]
  {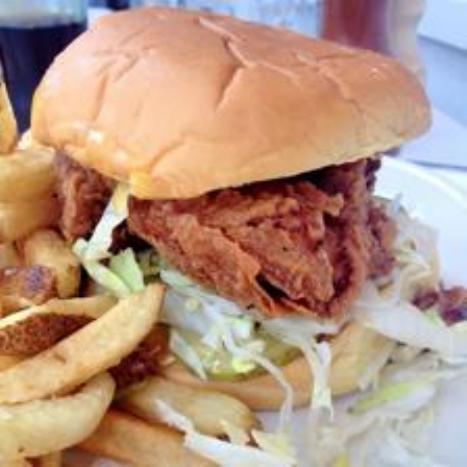}\\

  \includegraphics[width=0.14\textwidth]
  {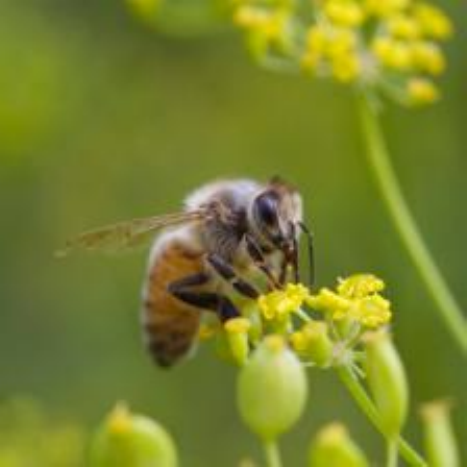}
  \includegraphics[width=0.14\textwidth]
  {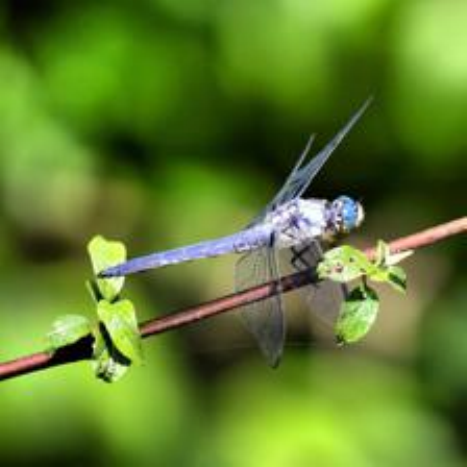}
  \includegraphics[width=0.14\textwidth]
  {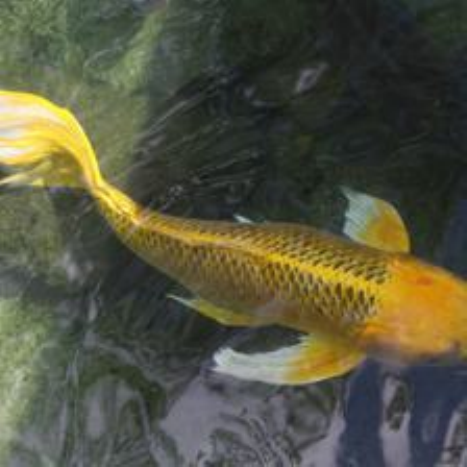}
  \includegraphics[width=0.14\textwidth]
  {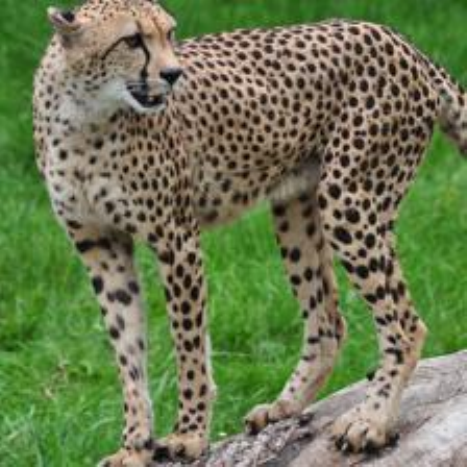}
  \includegraphics[width=0.14\textwidth]
  {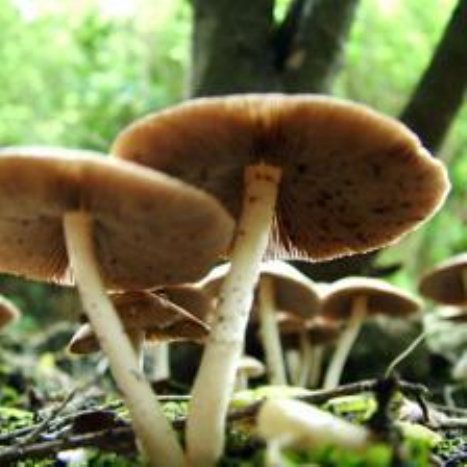}
  \includegraphics[width=0.14\textwidth]
  {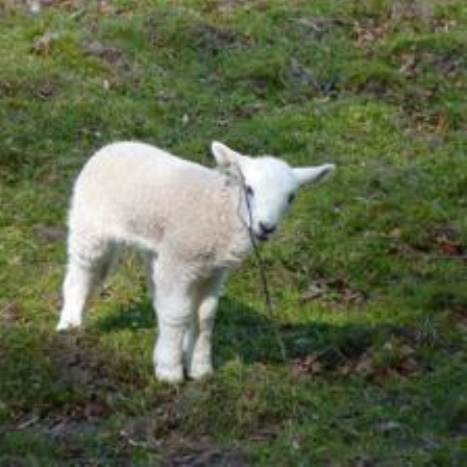}\\

  \includegraphics[width=0.14\textwidth]
  {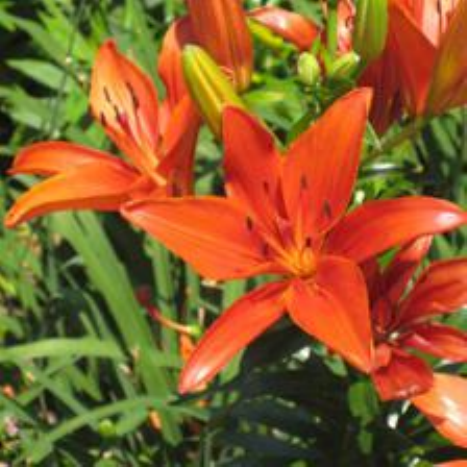}
  \includegraphics[width=0.14\textwidth]
  {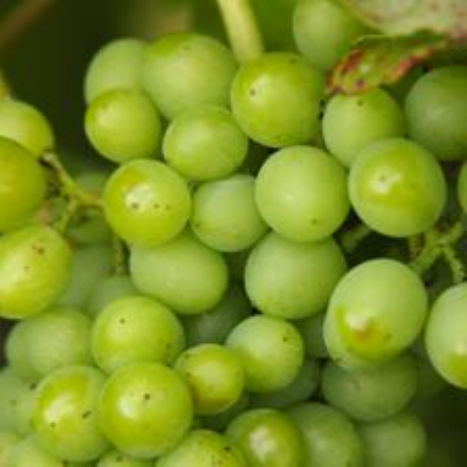}
  \includegraphics[width=0.14\textwidth]
  {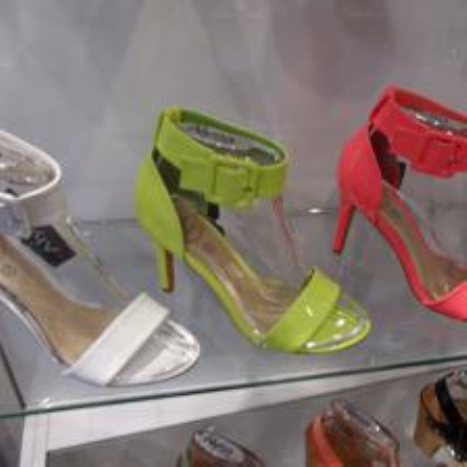}
  \includegraphics[width=0.14\textwidth]
  {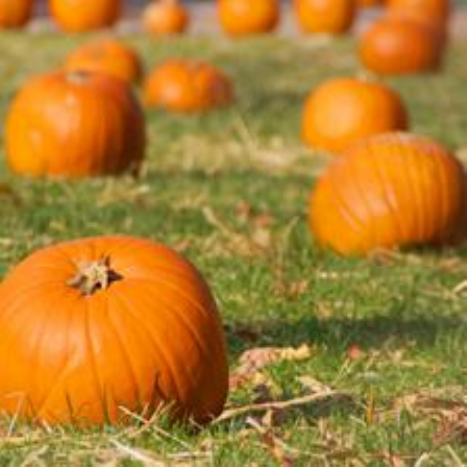}
  \includegraphics[width=0.14\textwidth]
  {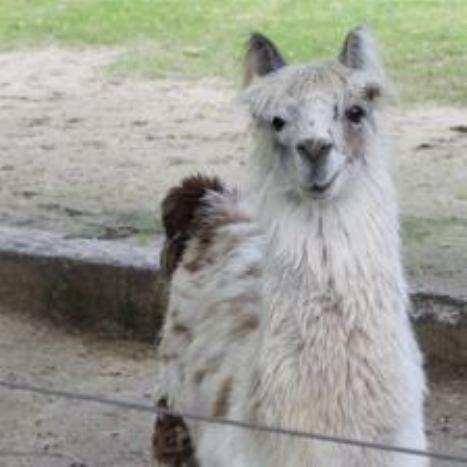}
  \includegraphics[width=0.14\textwidth]
  {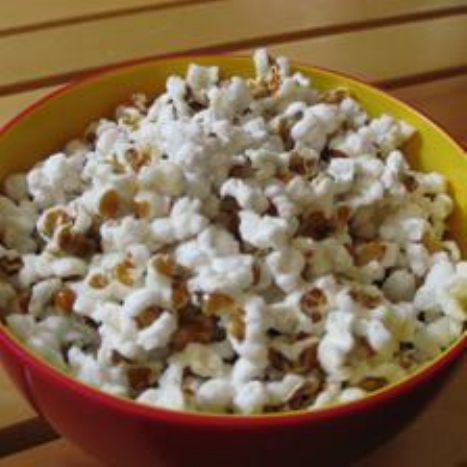}

  \caption{The original images correspond to Fig. \ref{fig:more_examples_3} and Fig \ref{fig:more_examples_4}. (first two rows) MS-COCO, and (the next two rows) Google Open Images.}
  \label{fig:originals_mscoco_goi}
\end{figure*}

Fig. \ref{fig:originals_imagenet}  provides the original images from ImageNet used in Fig. \ref{fig:more_examples} and Fig. \ref{fig:more_examples_2}. Similarly, Fig. \ref{fig:originals_mscoco_goi} shows the original images from MS-COCO and Open Images dataset used to generate the Fig. \ref{fig:more_examples_3} and Fig. \ref{fig:more_examples_4}.

\end{document}